\documentclass{article}

\usepackage[table,dvipsnames]{xcolor}
\usepackage[preprint]{corl_2026} %

\usepackage{graphicx}

\usepackage[utf8]{inputenc} %
\usepackage[T1]{fontenc}    %
\usepackage{hyperref}       %
\usepackage{url}            %
\usepackage{booktabs}       %
\usepackage{amsmath}
\usepackage{amsfonts}       %
\usepackage{nicefrac}       %
\usepackage{microtype}      %
\usepackage{cleveref}       %
\usepackage{lipsum}         %
\usepackage{graphicx}
\usepackage{natbib}

\usepackage{graphicx}
\usepackage{colortbl}
\usepackage{booktabs} 
\usepackage{xspace} 
\usepackage[raster,skins,most]{tcolorbox}
\usepackage{multirow}
\usepackage{arydshln}
\usepackage{makecell}
\usepackage{amssymb}
\usepackage{array}
\usepackage{enumitem}

\definecolor{rowgray}{gray}{0.96}
\definecolor{lightRed}{RGB}{252,243,242}
\definecolor{lightGreen}{RGB}{244,249,245}
\definecolor{lightOrange}{RGB}{253,246,239}
\definecolor{lightYellow}{RGB}{255,253,238}
\definecolor{lightPurple}{RGB}{243,243,253}
\definecolor{lightLime}{RGB}{245,255,243}
\definecolor{lightCream}{RGB}{255,250,240}
\definecolor{lightCyan}{RGB}{240,252,252}
\definecolor{lightPink}{RGB}{247,235,239}
\definecolor{lightPeach}{RGB}{255,245,240}

\definecolor{navyblue}{HTML}{0071BC}
\graphicspath{{assets/images/}}

\newcommand{\name}[0]{WatchAct\xspace}

\usepackage{pifont}
\definecolor{darkgreen}{RGB}{30, 150, 30}
\newcommand{\gcheck}{\textcolor{darkgreen}{\ding{51}}}
\newcommand{\rcross}{\textcolor{red}{\ding{55}}}

\usepackage{wrapfig}
\usepackage{graphicx}
\usepackage{subcaption}

\definecolor{citecolor}{HTML}{0071BC}

\hypersetup{
  colorlinks,
  linkcolor={black},
  citecolor={citecolor}
}

\graphicspath{{pics/}}

\title{WatchAct: A Benchmark for Behavior-Grounded Robot Manipulation}

\author{
  Baiqi Li\thanks{Corresponding author: \texttt{baiqili@cs.unc.edu}}
  \quad Ce Zhang
  \quad Yu Fang
  \quad Yue Yang
  \quad Shangzhe Li
  \quad Mingyu Ding
  \quad Gedas Bertasius\\
  \\
  University of North Carolina at Chapel Hill
}

\begin{document}
\maketitle

\begin{abstract}
A robot working alongside people must reason about what they have done, in what order, and with what intent. Video carries the spatial
layouts, object histories, and gestures that language leaves underspecified, yet today's manipulation benchmarks pair an instruction with a single current image, offering no way to evaluate
reasoning over observed human behavior. We introduce WatchAct, a
benchmark for robot manipulation grounded in observed human behavior.
Each instance pairs a real-world human-action video and a language
instruction with an aligned simulator scene and an executable LIBERO
task, enabling scalable and reproducible evaluation. WatchAct
comprises 3,000 long-horizon instances across 14 tasks in four
capability domains drawn from the cognitive demands of watching
another agent: parsing events (Event Grounding), recovering
procedural structure (Procedural Reasoning), inferring unstated
intent (Implicit Intent Inference), and tracking how the scene was
changed (Episodic Reasoning). We further propose a disentangled
evaluation protocol that separately measures (i)~video-to-plan
reasoning by vision-language models, (ii)~policy execution under
oracle plans, and (iii)~full task completion by integrated
planner--policy pipelines. In both simulation and on a Franka
Research 3 robot, current systems remain far from solving WatchAct.
The best pipeline, Gemini-3.1-Pro with $\pi_{0.5}$, reaches only
16.3\% Success Rate (SR) in simulation and 14.0\% on the real robot.
Gemini-3.1-Pro attains just 36.8\% Plan SR (vs.\ 97.1\% for
humans), while $\pi_{0.5}$ reaches only 21.5\% Task SR under oracle plans
and drops to 10.6\% on out-of-domain scenarios. Dataset and code are available at \url{https://baiqi-li.github.io/watchact_page/}.
\end{abstract}


\section{Introduction}
\vspace{-0.2cm}

A useful household or assistive robot spends most of its time alongside 
people~\citep{ciocarlie2012mobile, wu2023tidybot}. To act sensibly in such settings, it needs to make sense of 
what the people around it have done, what they were trying to 
accomplish, and what state they have left the scene in. Consider a 
kitchen in which a person opens a drawer, takes out several spice 
jars to cook with, and leaves them scattered on the counter. Told to 
``put everything back where it came from,'' a robot must reason about 
events it watched --- which jar came from which compartment, and in 
what arrangement. Acting on this instruction demands that the robot 
understand the person's actions and intent, not just the current scene.

Understanding other people's behavior is a longstanding object of study in cognitive 
science~\citep{sisbot2007human}. Humans can parse continuous observation into 
structured events with agents, content, and 
goals~\citep{zacks2007event}, infer unstated intentions from the 
actions~\citep{csibra2009natural, 
baker2009action}, and 
form persistent episodic representations of how others have changed 
a scene. These inferences rely on what video naturally captures --- 
spatial layouts, object histories, manual gestures, and procedural 
cues that language alone leaves underspecified. Robots collaborating 
with people need to draw similar inferences from the video they 
observe.

\begin{figure}[ht]
    \centering
    \includegraphics[width=0.9\linewidth]{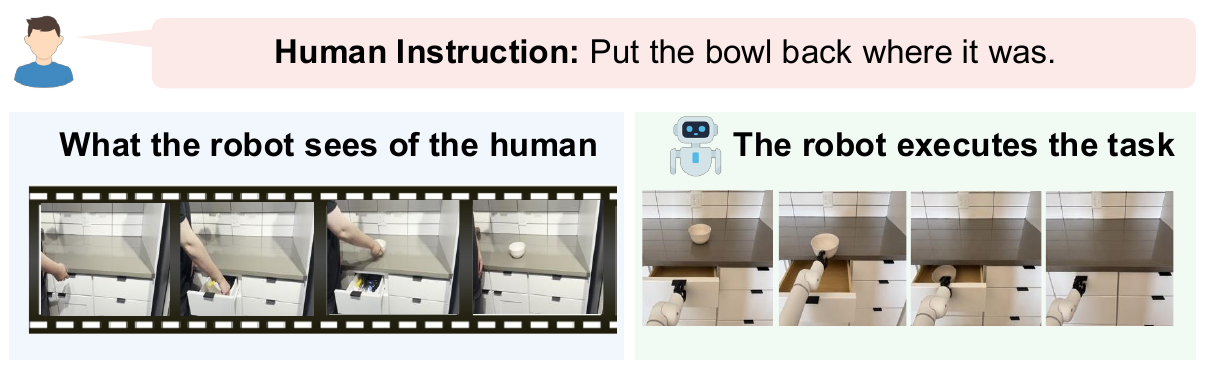}
    \caption{
    \textbf{WatchAct} is a behavior-grounded benchmark for robotic manipulation, where the robot reasons over observed human behavior and a language instruction to perform the corresponding task.\vspace{-0.6cm}
    }
    \label{fig:teaser}
\end{figure}

Despite the importance of such reasoning, existing manipulation benchmarks rarely test for it. Standard benchmarks evaluate execution from a static scene observation~\citep{liu2023libero, nasiriany2024robocasa, mees2022calvin, fei2025libero, zhou2025libero} and do not provide temporal context about prior events in the scene. Benchmarks that do incorporate video typically use it as a demonstration to be reproduced~\citep{jiang2023vima, wang2025vlm, son2026seetraceact} or as the robot's own self-view across past trials~\citep{dai2026robomme,hong2026esibench}, rather than as evidence about what another agent has done in the scene.

We introduce \name, a benchmark for robot manipulation grounded in 
observed human behavior. Each instance pairs a real-world 
human-action video and a language instruction with an aligned 
simulator scene and an executable task definition in LIBERO~\cite{liu2023libero}, enabling 
scalable and reproducible evaluation. \name comprises 3,000 
long-horizon instances across 14 tasks organized into four capability 
domains, each corresponding to a cognitive demand of watching another 
agent. \textit{Event Grounding} requires identifying task-relevant 
actions, states, and moments in the video. \textit{Procedural 
Reasoning} requires recovering the temporal structure of observed 
action sequences. \textit{Implicit Intent Inference} requires 
inferring unstated goals from gestures and ambiguous references. 
\textit{Episodic Reasoning} requires tracking how the scene was 
changed and supporting higher-level planning that depends on this 
history. Across these tasks, \name spans long-horizon planning, fine-grained action understanding, spatial reasoning, and inference over human goals. We further vary camera viewpoint and spatial reference frame across instances to test spatial generalization.

We evaluate current systems under a disentangled protocol that 
separately measures video-to-plan reasoning, oracle-plan execution, 
and integrated pipeline performance. Each stage exposes a substantial 
failure mode. The strongest VLM, Gemini-3.1-Pro~\citep{google2026gemini31pro}, reaches only 36.8\% planning 
success, 60.3\% below human performance, and falters on 
human-centered references and long horizons. Even when given oracle 
plans, $\pi_{0.5}$ achieves 21.5\% overall, indicating that 
reliable execution is itself an open problem. When the two stages are combined into a pipeline, their errors compound to 16.3\% success. 
Real-robot experiments on a Franka Research 3~\citep{franka2026research3} reproduce these patterns, with the integrated Gemini-3.1-Pro and $\pi_{0.5}$ pipeline reaching only 14.0\%.

We make three contributions. \textbf{(1)~Benchmark.} \name, the first
benchmark to evaluate robot manipulation grounded in observed human
behavior, comprising 3,000 long-horizon instances across 14 tasks in
four cognitively-motivated capability domains, each pairing a
real-world human-action video with an aligned, executable LIBERO
simulator task. \textbf{(2)~Disentangled evaluation.} A protocol that
separately measures VLM video-to-plan reasoning, policy execution
under oracle plans, and integrated planner--policy performance,
exposing where errors originate along the perception-to-action chain.
\textbf{(3)~Findings.} Evaluating 9 VLMs and 4 policies in simulation
and on a Franka Research 3 robot, we identify human-centered spatial
reasoning, long-horizon planning, and physical execution as the main
sources of failure for current systems.

\vspace{-0.2cm}
\section{Related Work}
\label{sec:related-work}
\vspace{-0.2cm}

\begin{table*}[t]
\centering
\small
\renewcommand{\arraystretch}{1.3}
\setlength{\tabcolsep}{5pt}
\vspace{-0.3cm}
\caption{\textbf{Comparison with prior manipulation benchmarks.}
\textsc{Capabilities} columns (Event, Procedural, Intent, Episodic)
mark coverage of the four cognitive domains from
Section~\ref{sec:benchmark}. \textsc{Plan} and \textsc{Exec.} indicate
whether the benchmark evaluates planning and low-level execution.}
\label{tab:related_work_comparison}
\begin{tabular}{l c c c c c c}
\toprule
\textsc{\textbf{Benchmark}} &
\multicolumn{4}{c}{\textsc{\textbf{Capabilities}}} &
\multicolumn{2}{c}{\textsc{\textbf{Evaluation}}} \\
\cmidrule(lr){2-5} \cmidrule(lr){6-7}
& \textsc{Event} & \textsc{Procedural} & \textsc{Intent} & \textsc{Episodic} & \textsc{Plan} & \textsc{Exec.} \\
\midrule
LIBERO~\cite{liu2023libero}             & \rcross & \rcross & \rcross & \rcross & \rcross & \gcheck \\
RoboCasa~\cite{nasiriany2024robocasa}   & \rcross & \rcross & \rcross & \rcross & \rcross & \gcheck \\
CALVIN~\cite{mees2022calvin}            & \rcross & \rcross & \rcross & \rcross & \rcross & \gcheck \\
BEHAVIOR-1K~\cite{li2023behavior}       & \rcross & \rcross & \rcross & \rcross & \rcross & \gcheck \\
VLABench~\cite{zhang2025vlabench}       & \rcross & \rcross & \rcross & \rcross & \gcheck & \gcheck \\
RoboMME~\cite{dai2026robomme}           & \rcross & \gcheck & \rcross & \gcheck & \rcross & \gcheck \\
MIKASA-Robo~\cite{cherepanov2025mikasa} & \rcross & \gcheck & \rcross & \gcheck & \rcross & \gcheck \\
RoboCerebra~\cite{han2025robocerebra}   & \rcross & \gcheck & \rcross & \gcheck & \gcheck & \gcheck \\
ESI-Bench~\cite{hong2026esibench}       & \gcheck & \rcross & \rcross & \rcross & \gcheck & \rcross \\
VIMA-Bench~\cite{jiang2023vima}         & \rcross & \gcheck & \rcross & \rcross & \rcross & \gcheck \\
\midrule
\textbf{Ours} & \gcheck & \gcheck & \gcheck & \gcheck & \gcheck & \gcheck \\
\bottomrule
\end{tabular}
\end{table*}

\noindent\textbf{Manipulation Benchmarks with Temporal or Video Context.}
Manipulation benchmarks differ in what temporal context, if any, the
robot is given at test time, and we organize prior work by its source.
Standard benchmarks pair a language instruction with the current scene
observation and evaluate execution from this static input
alone~\citep{liu2023libero, nasiriany2024robocasa, shridhar2020alfred,
james2020rlbench, zhang2025vlabench, mees2022calvin, li2023behavior}.
A second line treats the robot's own prior trajectory as the temporal
context, evaluating memory over the agent's self-view across past
trials~\citep{dai2026robomme, cherepanov2025mikasa, chen2026rmbench,
fang2025sam2act, han2025robocerebra}. ESI-Bench~\citep{hong2026esibench} goes further by asking the agent to
actively gather that context. A third line introduces video as a
demonstration to reproduce, including one-shot video
imitation~\citep{jiang2023vima, li2024okami, heppert2024ditto, park2025demodiffusion}, VLM planning from human pick-and-place
demos~\citep{wang2025vlm}, and policy learning from human play
videos~\citep{wang2023mimicplay, pan2025mimicdroid}. In \name, the robot reasons about another person's activity in the scene, using the video as evidence about object histories, spatial relations, and intent rather than as a trajectory to reproduce or a memory to recall.

\noindent\textbf{VLM and VLA Architectures for Manipulation.}
End-to-end vision-language-action models such as
$\pi_{0.5}$~\citep{intelligence2025pi_} and
OpenVLA~\citep{kim2024openvla} map observations and instructions
directly to actions. A complementary family decouples high-level
reasoning from low-level control by pairing a VLM planner with a
downstream policy~\citep{wang2025vlm, shi2025hi, li2025hamster,
zhang2025hirt,yang2026seeing}. \name evaluates representative members of both families
under a disentangled protocol that separately measures video-to-plan
reasoning, policy execution under oracle plans, and end-to-end
completion by integrated pipelines.

\vspace{-0.3cm}
\section{The \name Benchmark}
\label{sec:benchmark}
\vspace{-0.2cm}

\name evaluates robot manipulation grounded in observed human behavior.
Each instance pairs a real-world video of a human acting in a scene
with a language instruction and an aligned, executable simulator task. We formalize the problem (Section~\ref{sec:formulation}), introduce
the task taxonomy (Section~\ref{sec:taxonomy}), describe instance
construction (Section~\ref{sec:construction}), and detail the spatial
conditions we vary (Section~\ref{sec:spatial_design}).

\vspace{-0.3cm}
\subsection{Problem Formulation}
\label{sec:formulation}
\vspace{-0.1cm}

We consider video-grounded robot manipulation, where an agent acts on
a language instruction given a video of human activity in the scene
and the current scene state. Formally, each benchmark instance is defined as
\(\mathcal{I} = (V, \ell, \mathcal{T})\), where \(V\) is a real-world
video, \(\ell\) is a language instruction, and \(\mathcal{T}\) is an
executable embodied task. The task \(\mathcal{T}\) specifies the
execution environment, the initial state \(x^0\), the observation
space \(\mathcal{O}\), the low-level action space \(\mathcal{A}\), and
a success predicate \(\phi\). The video-language context
\(c=(V,\ell)\) encodes latent task variables such as relevant object
identities, prior object locations, action order, spatial reference
frames, and human intent. The agent starts from \(x^0\) and at each
timestep receives a partial observation \(o_t \in \mathcal{O}\) and the
context \(c\), then outputs an action \(a_t \in \mathcal{A}\) until
\(\phi\) is satisfied or the maximum episode length is reached.

Each instance is also annotated with an oracle action plan
\(P^\star = (u_1^\star, \ldots, u_K^\star)\) over the high-level
primitive actions \(\{\texttt{Pick}, \texttt{Place},
\texttt{Open}, \texttt{Close}\}\). This annotation supports the
disentangled evaluation protocol in Section~\ref{sec:experiments},
which separates video-to-plan reasoning from low-level policy
execution.

\vspace{-0.6cm}
\subsection{Task Taxonomy}
\label{sec:taxonomy}
\vspace{-0.1cm}

\begin{table*}[t]
\centering
\small
\setlength{\tabcolsep}{3pt}
\renewcommand{\arraystretch}{1.2}
\caption{\textbf{Task taxonomy.} \name comprises 14 tasks
grouped into four cognitive capabilities of observing another agent. The Count column reports the number of
instances per task.}
\label{tab:taxonomy_brief}
\resizebox{\textwidth}{!}{
\begin{tabular}{l p{12.7cm} c}
\toprule
\textbf{Task Name} & \textbf{Brief Description} & \textbf{Count} \\
\midrule
\rowcolor{lightOrange}
\multicolumn{3}{c}{\textbf{Event Grounding}} \\
Fine-Grained Action & The robot uses fine-grained observed actions to infer and execute the task. & 225 \\
\addlinespace[2pt]
\rowcolor{rowgray}
Count & The robot infers and executes the task by counting how many times relevant events occur in the video. & 160 \\
\addlinespace[2pt]
Ordinal & The robot grounds the task in a specified ordinal event. & 215 \\
\addlinespace[2pt]
\rowcolor{rowgray}
State Change & The robot infers and executes the task by tracking state changes in the scene. & 155 \\
\addlinespace[2pt]
Moment & The robot localizes a described event in the video and retrieves the relevant information for reasoning. & 135 \\
\addlinespace[2pt]
\midrule
\rowcolor{lightRed}
\multicolumn{3}{c}{\textbf{Procedural Reasoning}} \\
Imitation & The robot reproduces the same action sequence performed by the human. & 290 \\
\addlinespace[2pt]
\rowcolor{rowgray}
Reversal & The robot reverses the human's action sequence to undo its effects. & 225 \\
\addlinespace[2pt]
Temporal Sort & The robot spatially arranges objects according to the event order. & 210 \\
\addlinespace[2pt]
\midrule
\rowcolor{lightYellow}
\multicolumn{3}{c}{\textbf{Implicit Intent Inference}} \\
Nonverbal Cue & The robot infers the intended task from human nonverbal cues, such as gestures or hand motions. & 195 \\
\addlinespace[2pt]
\rowcolor{rowgray}
Reference Disambiguation & The robot resolves ambiguous language references using the observed video context. & 260 \\
\addlinespace[2pt]
\midrule
\rowcolor{lightCream}
\multicolumn{3}{c}{\textbf{Episodic Reasoning}} \\
Restore Previous State & The robot returns displaced objects to their initial positions. & 285 \\
\addlinespace[2pt]
\rowcolor{rowgray}
Task Continuation & Given a task instruction, the robot completes the unfinished parts of the task. & 170 \\
\addlinespace[2pt]
Error Correction & Given a task instruction, the robot detects and corrects errors in the human action video. & 240 \\
\addlinespace[2pt]
\rowcolor{rowgray}
Conditional Execution & The robot conditionally executes different actions based on events observed in the video. & 235 \\
\bottomrule
\end{tabular}}
\end{table*}

Humans observing other humans do not process visual input as an
unstructured stream. They parse it into meaningful events along
multiple dimensions that matter for acting in shared
environments---the content of each
event~\citep{kurby2008segmentation, zacks2007event}, the intent
behind it~\citep{abowd1999towards, sebanz2006joint}, its temporal
relations to other events~\citep{zwaan1995construction}, and its
causal consequences~\citep{schank2013scripts}. A robot reasoning
about a person's activity in a shared scene faces the same cognitive
demands. It must identify what happened, infer why, recover when, and
track the resulting state of the world. \name organizes its 14 task
scenarios around these four cognitive demands, grouped into four
capability domains.

\textbf{Event Grounding} requires identifying a task-relevant event
by its action, state, or moment and acting on it accordingly.
\textbf{Procedural Reasoning} requires understanding the temporal
sequence or structure of observed events and acting with this
procedural information. \textbf{Implicit Intent Inference} requires
inferring unstated human intent from cues such as gestures, hand
motions, or ambiguous references. \textbf{Episodic Reasoning}
requires reasoning over event histories to support higher-level
planning, such as continuing unfinished tasks or correcting errors.
Table~\ref{tab:taxonomy_brief} summarizes the 14 task scenarios, their
distribution across the four domains, and the number of instances per
task.

\vspace{-0.2cm}
\subsection{Benchmark Construction}
\label{sec:construction}
\begin{figure}[ht]
    \centering
    \includegraphics[width=\linewidth]{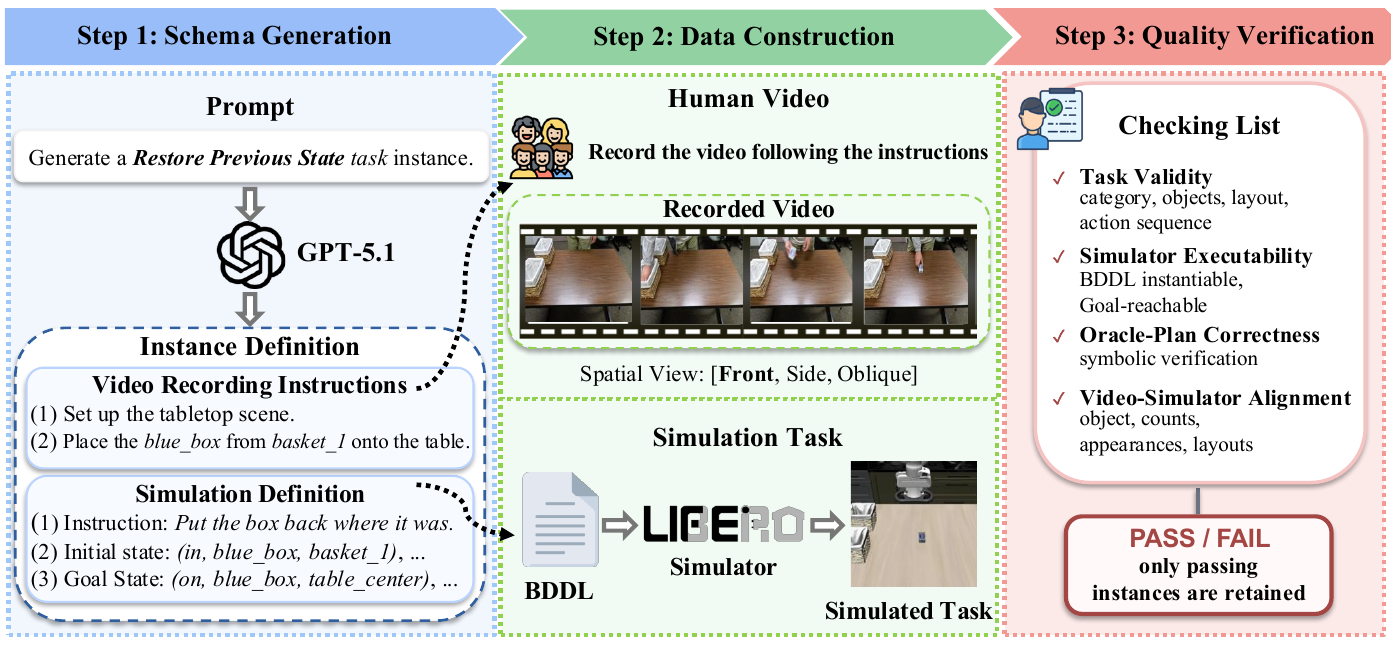}
\caption{
\textbf{WatchAct data construction pipeline.} \textit{Step 1 (Schema Generation):} GPT-5.1 generates video recording instructions and the corresponding simulation task definition. \textit{Step 2 (Data Construction):} we record the human video following the instructions and instantiate the matching LIBERO task. \textit{Step 3 (Quality Verification):} each instance passes four checks: (i)~Task Validity, (ii)~Simulator Executability, (iii)~Oracle-Plan Correctness, and (iv)~Video--Simulator Alignment.\vspace{-0.5cm}}
    \label{fig:data_pipeline}
\end{figure}

Each \name instance pairs a real-world video of human activity with
an aligned, executable simulator task. The alignment must hold along
three dimensions. \textbf{Execution-state consistency}: the
simulator's initial state matches the moment in the video from which
the robot is expected to act. \textbf{Scene consistency}: objects in
the simulator match the video objects in category, appearance, and
relative size, with layouts aligned through a shared set of named
regions. \textbf{Goal consistency}: the simulator's success predicate
matches the goal implied by the video and instruction. We construct \name through a three-stage pipeline illustrated in Figure~\ref{fig:data_pipeline}. 

\noindent\textbf{Stage 1: Task Specification.} We prompt a frontier
LLM (e.g., GPT-5.1~\citep{openai2026gpt51}) with a shared inventory of objects and named
regions, registered between the simulator and the real-world scene,
along with the capability taxonomy from
Table~\ref{tab:taxonomy_brief}. The LLM produces a complete task
specification comprising four artifacts: a language instruction, a
filming guide for the human demonstrator, a LIBERO BDDL task
definition~\citep{liu2023libero} that fixes the simulator's initial scene state and success
predicate, and an oracle high-level action plan over the primitive
action space.

\noindent\textbf{Stage 2: Video Recording.} Following the filming
guide, we instantiate the real-world scene with the specified objects
and layout, and a human demonstrator records the action sequence
(e.g., \textit{``Step 1: pick up the milk. Step 2: place it in the
basket near the human''}). Multiple takes are recorded per instance,
and we retain the take that most clearly follows the filming guide.
Videos are filmed from one of several camera angles to support the
spatial variation described in Section~\ref{sec:spatial_design}.

\noindent\textbf{Stage 3: Simulator Instantiation.} The BDDL task
definition is instantiated in LIBERO~\citep{liu2023libero}. We extend LIBERO with
additional real-world objects obtained through 3D scanning to match
the objects used in the video, and we define new success predicates
required by the four capability domains. Object layouts in the
simulator are aligned with the video through the shared named
regions---11 tabletop regions (e.g., \textit{front-left},
\textit{front-right}) and 6 container regions (e.g.,
\textit{inside-basket}). The initial simulator state corresponds to
the moment in the video from which the robot is expected to act: for
most tasks, the final state of the video (so the robot continues from
the outcome of the observed activity), and for imitation tasks, the
state before the human's action (so the robot reproduces the sequence
from the same initial conditions).

\noindent\textbf{Stage 4: Quality Verification.} Each instance is
manually verified along four axes.
\textit{Task validity}---the generated task matches the intended
category, uses the specified objects, and contains plausible layouts
and action sequences.
\textit{Simulator executability}---the BDDL task instantiates in
LIBERO, all required interactions are feasible, and a teleoperated
solution reaches the success predicate.
\textit{Oracle-plan correctness}---the oracle plan is run through a
symbolic executor that simulates its effect on the scene, and the
resulting object states must satisfy the success predicate. Only
instances that pass are retained.
\textit{Video--simulator alignment}---the recorded video matches the
simulator scene in object categories, counts, appearances, and
layouts, and satisfies the execution-state, scene, and goal
consistency dimensions stated at the start of this section.

\subsection{Spatial Reasoning Design}
\label{sec:spatial_design}
\vspace{-0.2cm}

When recording human video, we vary spatial conditions along two axes. First, videos are captured from diverse \textbf{camera
viewpoints}, including frontal, side, and oblique (rear- and
front-oblique) views. Because the robot's simulator observation is
fixed, the model must reason about spatial relations that hold
regardless of the angle the video was filmed from. Second,
language instructions use varied \textbf{reference frames}, either
human-centered (``the cup on the human's left'') or camera-centered.
Together these axes test whether agents can ground spatial relations
observed from one perspective, interpret frame-dependent spatial
language, and act from the robot's execution viewpoint.

\vspace{-0.3cm}
\section{Experiments}
\label{sec:experiments}
\vspace{-0.2cm}

\subsection{Experimental Setup}
\vspace{-0.2cm}

\noindent\textbf{Evaluation Protocol.}
\name evaluates video-grounded embodied agents along three dimensions:
(i) video-to-plan reasoning, (ii) policy execution with oracle plans, and
(iii) full task completion by integrated planner and policy pipelines. We instantiate the planner with a VLM, which provides strong video-to-plan reasoning ability.
\begin{itemize}[leftmargin=*, itemsep=0pt, topsep=0pt]
    \item \textit{Video-to-plan Reasoning.} Given a video and a language instruction, the planner predicts a high-level task plan. To evaluate the plan, we run it through a symbolic executor that simulates its effect on the scene, and check whether the resulting object states satisfy the task's success conditions. Details are provided in the Appendix.
    \item \textit{Policy Execution with Oracle Plans.} The policy is given the ground-truth oracle plan (a correct sequence of high-level commands) and executes it one command at a time, with each command held for a fixed number of steps before advancing to the next. This setting isolates the policy's ability to carry out a correct plan, independent of any planning errors.
    \item \textit{Integrated Planner and Policy.} The agent conditions on the video, language instruction, and robot observation, and directly outputs low-level actions. Performance is measured by whether the robot reaches the task's success conditions before timing out.
\end{itemize}

\noindent\textbf{Metrics.} \textit{Plan SR} scores video-to-plan reasoning: whether a predicted high-level plan satisfies the task's success conditions under symbolic execution, with no robot involved. \textit{Task SR} scores physical execution: whether the robot reaches the success conditions in the simulator. We also report \textit{Progress Rate} (PR), the percentage of completed subgoals during execution. %

\noindent\textbf{Baselines.}
For video-to-plan reasoning, we evaluate representative open-source and commercial VLMs, including InternVL3.5-241B~\cite{wang2025internvl3}, Qwen3-VL-235B-Instruct/Thinking~\cite{bai2025qwen3}, GPT-5.4~\citep{openai2026gpt51}, Gemini-3.1-Pro~\citep{google2026gemini31pro}, and Gemini-Robotics-ER-1.6~\cite{graesser2026geminiroboticser16}.
For oracle-plan execution, we evaluate four state-of-the-art policies: three VLA models ($\pi_{0.5}$~\cite{intelligence2025pi_}, OpenVLA-OFT~\cite{kim2025fine}, UniVLA~\cite{bu2025univla}) and one world-model-based policy (LingBot-VA~\cite{li2026causal}).
For the integrated planner and policy setting, we pair GPT-5.4 or Gemini-3.1-Pro with each execution policy: the planner generates a multi-step plan, and the policy executes each step for a fixed control horizon.

\noindent\textbf{Implementation Details.}
For all VLMs, we sample videos from \name{} at 1 FPS and use each model's default inference parameters.
For simulated experiments involving robot policy execution, we run simulations on the LIBERO~\cite{liu2023libero} platform and evaluate each task with 10 trials. Each command is forwarded to the policy every 300 environment steps. All robotic policies use versions fine-tuned on LIBERO demonstrations. Full experimental details, including prompts and per-task configurations, are provided in the Appendix. Details of the real-world robot experiments are provided in Section~\ref{sec:real_robot}.

\begin{table}[t]
    \centering
 \caption{\textbf{Main Results on Video-to-plan Reasoning.} Each predicted VLM plan is run through a symbolic executor and scored using the Plan SR metric, which measures whether the resulting object states satisfy the task's success conditions. Even the strongest model, Gemini-3.1-Pro, reaches only 36.8\% average Plan SR, 60.3 points below human performance. Task abbreviations: \textbf{Act}: Action; \textbf{Cnt}: Count; \textbf{Ord}: Ordinal; \textbf{SC}: State Change; \textbf{Mmt}: Moment; \textbf{Imit}: Imitation; \textbf{Rev}: Reversal; \textbf{TS}: Temporal Sort; \textbf{NC}: Nonverbal Cue; \textbf{RD}: Reference Disambiguation; \textbf{Rest}: Restore; \textbf{TC}: Task Continuation; \textbf{EC}: Error Correction; \textbf{Cond}: Conditional Execution. See Table~\ref{tab:taxonomy_brief} for definitions.}
 \label{tab:subtask_results_vlm}
 \vspace{0.5em}
 \renewcommand{\arraystretch}{1.3}
 \setlength{\tabcolsep}{2pt}
 \resizebox{1\textwidth}{!}{
 \begin{tabular}{l|cccccc|cccc|ccc|ccccc|c}
 \toprule
 \multirow{3}{*}{Methods} &
 \multicolumn{6}{c}{\cellcolor{yellow!10}Event} &
 \multicolumn{4}{c}{\cellcolor{orange!10}Procedural} &
 \multicolumn{3}{c}{\cellcolor{pink!20}Implicit Intent} &
 \multicolumn{5}{c}{\cellcolor{brown!10}Episodic} &
 \multirow{3}{*}{Overall} \\
 &
 \multicolumn{6}{c}{\cellcolor{yellow!10}Grounding} &
 \multicolumn{4}{c}{\cellcolor{orange!10}Reasoning} &
 \multicolumn{3}{c}{\cellcolor{pink!20}Inference} &
 \multicolumn{5}{c}{\cellcolor{brown!10}Reasoning} & \\
 \cmidrule(lr){2-7} \cmidrule(lr){8-11} \cmidrule(lr){12-14} \cmidrule(lr){15-19}
 &
 Act & Cnt & Ord & SC & Mmt & \textbf{Avg.} &
 Imit & Rev & TS & \textbf{Avg.} &
 NC & RD & \textbf{Avg.} &
 Rest & TC & EC & Cond & \textbf{Avg.} & \\
 \midrule
 Human Performance
 & 100.0 & 93.3 & 93.3 & 100.0 & 100.0 & 97.3
 & 96.7 & 96.7 & 96.7 & 96.7
 & 93.3 & 96.7 & 95.0
 & 100.0 & 96.7 & 93.3 & 100.0 & 98.0
 & 97.1 \\
 \midrule
 \rowcolor{navyblue!5} \multicolumn{1}{l|}{\textcolor{black}{\textbf{\textit{Open-source Models}}}} & & & & & & & & & & & & & & & & & & & \\
 Qwen3-VL-235B-Instruct
 & 4.9 & 10.6 & 3.7 & 14.8 & 5.9 & 7.5
 & 12.8 & 8.4 & 8.6 & 10.2
 & 6.7 & 4.2 & 5.3
 & 5.0 & 27.6 & 22.4 & 18.3 & 17.0
 & 10.8 \\
 InternVL3.5-241B
 & 16.9 & 15.0 & 12.1 & 21.3 & 0.8 & 13.8
 & 7.3 & 4.0 & 4.7 & 5.5
 & 2.6 & 10.4 & 7.0
 & 4.2 & 33.5 & 30.4 & 22.7 & 21.0
 & 13.0 \\
 Qwen3-VL-235B-Thinking
 & 18.2 & 21.9 & 26.0 & 31.0 & 12.1 & 22.1
 & 9.7 & 6.7 & 7.2 & 8.0
 & 6.2 & 17.3 & 12.6
 & 10.8 & 44.1 & 36.0 & 30.3 & 28.3
 & 18.4 \\
 \midrule
 \rowcolor{navyblue!5} \multicolumn{1}{l|}{\textcolor{black}{\textbf{\textit{Proprietary Models}}}} & & & & & & & & & & & & & & & & & & & \\
 GPT-5.4
 & \textbf{36.4} & 24.4 & 33.0 & \textbf{60.6} & 4.5 & 32.9
 & \textbf{32.8} & 27.1 & 27.6 & \textbf{29.6}
 & \textbf{21.0} & 20.4 & 20.7
 & \textbf{36.9} & 43.5 & 37.6 & 39.3 & 39.1
 & 31.5 \\
 Gemini-Robotics-ER-1.6
 & 33.3 & \textbf{40.6} & 44.7 & 45.2 & 19.2 & 37.4
 & 23.1 & 25.3 & 22.2 & 23.5
 & 17.9 & 28.5 & 24.0
 & 33.5 & 48.2 & 37.6 & 39.8 & 38.9
 & 32.4 \\
 Gemini-3.1-Pro
 & 33.8 & 36.9 & \textbf{47.4} & 60.0 & \textbf{22.7} & \textbf{40.7}
 & 30.0 & \textbf{30.2} & \textbf{28.1} & 29.5
 & 18.0 & \textbf{43.1} & \textbf{32.4}
 & 31.2 & \textbf{52.4} & \textbf{43.2} & \textbf{42.3} & \textbf{41.0}
 & \textbf{36.8} \\
 \bottomrule
 \end{tabular}}
 \end{table}

\vspace{-0.2cm}
\subsection{Main Results}

We report main results on video-to-plan reasoning in Table~\ref{tab:subtask_results_vlm}, broken down by the 14 task categories (Table~\ref{tab:taxonomy_brief}), and on oracle-plan execution and the integrated pipeline in Table~\ref{tab:action_plan_execution}.

\begin{table}[t]
\small
\centering
\caption{ \textbf{Policy Execution Performance.} We evaluate robotic policies under three plan sources: oracle plans and plans produced by GPT-5.4 and Gemini-3.1-Pro. Policies struggle on \name even with oracle plans: the best performer, $\pi_{0.5}$, reaches only 21.5 Task SR and 33.0 PR.} %
\label{tab:action_plan_execution}
\setlength{\tabcolsep}{10pt}
\renewcommand{\arraystretch}{1.1}
\begin{tabular}{l|cc|cc|cc}
\toprule
Method & \multicolumn{2}{c|}{\cellcolor{pink!20}Oracle Plan} &
\multicolumn{2}{c|}{\cellcolor{yellow!10}GPT-5.4} & \multicolumn{2}{c}{\cellcolor{orange!10}Gemini-3.1-Pro}
\\ & Task SR & PR & Task SR & PR & Task SR & PR
\\ \midrule
Oracle Execution & 100.0 & 100.0 & 31.5 & 40.1 & 36.8 & 45.5 \\
\midrule
$\pi_{0.5}$ & \textbf{21.5} & \textbf{33.0} & \textbf{13.5} & \textbf{21.5} & \textbf{16.3} & \textbf{23.0} \\
OpenVLA-oft & 1.1 & 1.6 & 0.2 & 0.9 & 0.3 & 0.9 \\
UniVLA & 0.6 & 1.3 & 0.0 & 0.5 & 0.0 & 1.3 \\
Lingbot-VA & 0.9 & 2.7 & 0.3 & 1.2 & 0.4 & 2.1 \\
\bottomrule
\end{tabular}
\end{table}

\noindent\textbf{VLMs struggle with video-to-plan reasoning.}
All models perform poorly on video-to-plan reasoning (Table~\ref{tab:subtask_results_vlm}). To calibrate this difficulty against human ability, we ask four annotators to produce high-level plans from the same video--language inputs and score them with the same symbolic protocol used for VLMs. Annotators reach 97.1\% Plan SR, while the best model, Gemini-3.1-Pro, reaches only 36.8\%, a gap of 60.3 points. This indicates that inferring a valid manipulation plan from video remains a major challenge for current VLMs. We also observe a large gap between open- and closed-source models. Averaged across categories, the strongest open-source model, Qwen3-VL-235B-Thinking, reaches 18.4\% Plan SR, trailing Gemini-3.1-Pro by 18.4 points.

\noindent\textbf{Robotic policies struggle to follow oracle plans.}
Table~\ref{tab:action_plan_execution} reports Task SR under oracle plans, which measures each policy's ability to execute a correct high-level plan and provides an upper bound for the corresponding integrated pipeline. Even with oracle plans, current policies struggle on \name. The best policy, $\pi_{0.5}$, reaches only 21.5\% Task SR, far from solving the long-horizon tasks in the benchmark, and the remaining policies fall below 2\%.

\noindent\textbf{Errors accumulate across the pipeline.}
Table~\ref{tab:action_plan_execution} also reports the integrated setting. Predicted plans degrade execution relative to oracle plans. Paired with $\pi_{0.5}$, Gemini-3.1-Pro plans reach 16.3\% Task SR and GPT-5.4 plans 13.5\%, both below the 21.5\% oracle-plan ceiling.

\vspace{-0.2cm}
\subsection{Diagnostic Analysis}
\vspace{-0.1cm}

\noindent\textbf{Long-horizon Scaling.}
We next analyze how plan length, the number of high-level actions in a
task's oracle plan, affects reasoning and execution. As plan length
grows from 2 to 6 actions, GPT-5.4's Plan SR drops from 36.7\% to
7.2\% and $\pi_{0.5}$'s oracle-plan Task SR from 37.3\% to 0.0\%. The
decline holds across all VLMs and both plan sources, indicating that robust
long-horizon behavior remains challenging for both planning and execution.

\begin{figure*}[t]
    \centering
    \begin{minipage}[t]{0.32\textwidth}
        \centering
        \includegraphics[width=\linewidth]{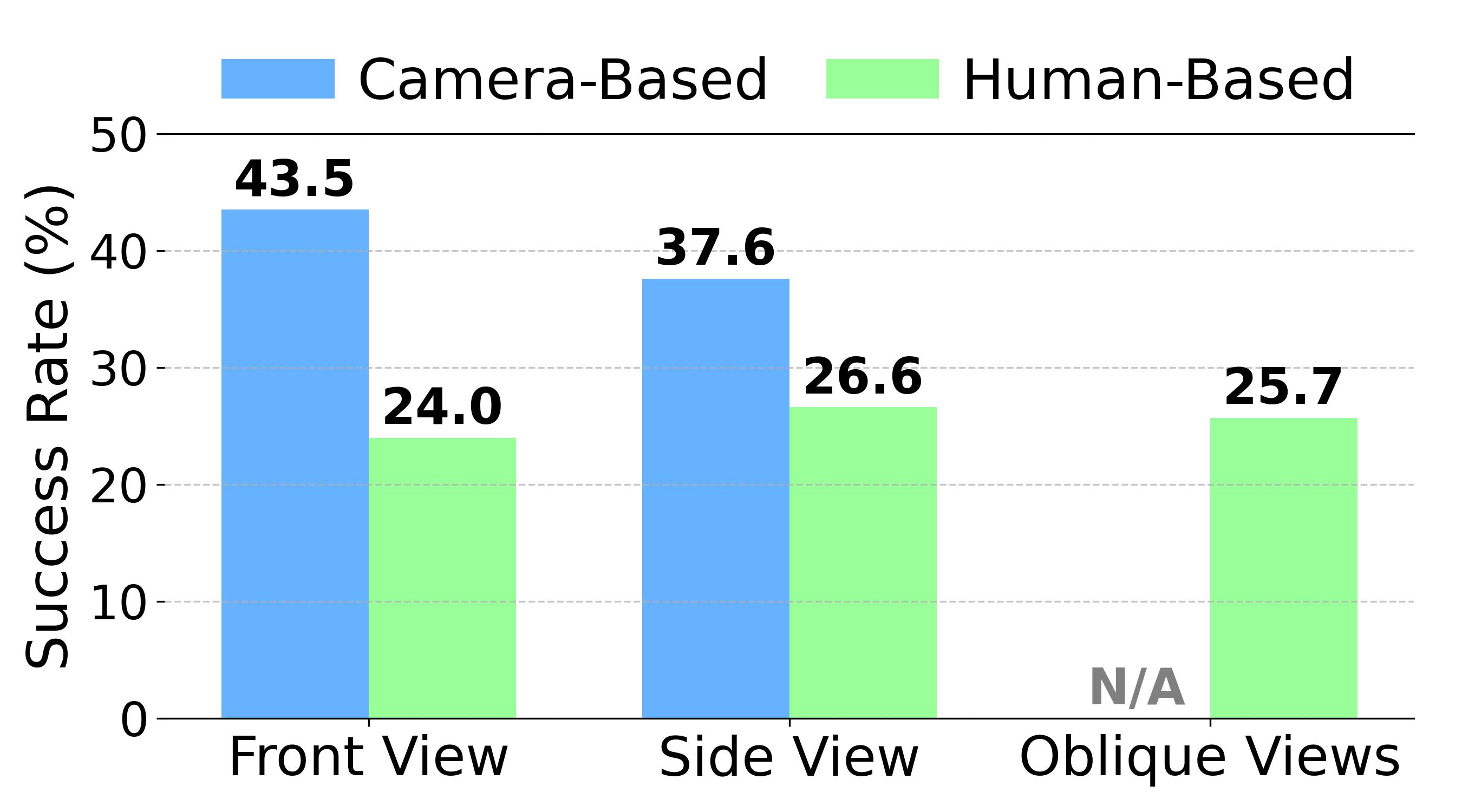}
        {\small (a) GPT-5.4}
    \end{minipage}
    \hfill
    \begin{minipage}[t]{0.32\textwidth}
        \centering
        \includegraphics[width=\linewidth]{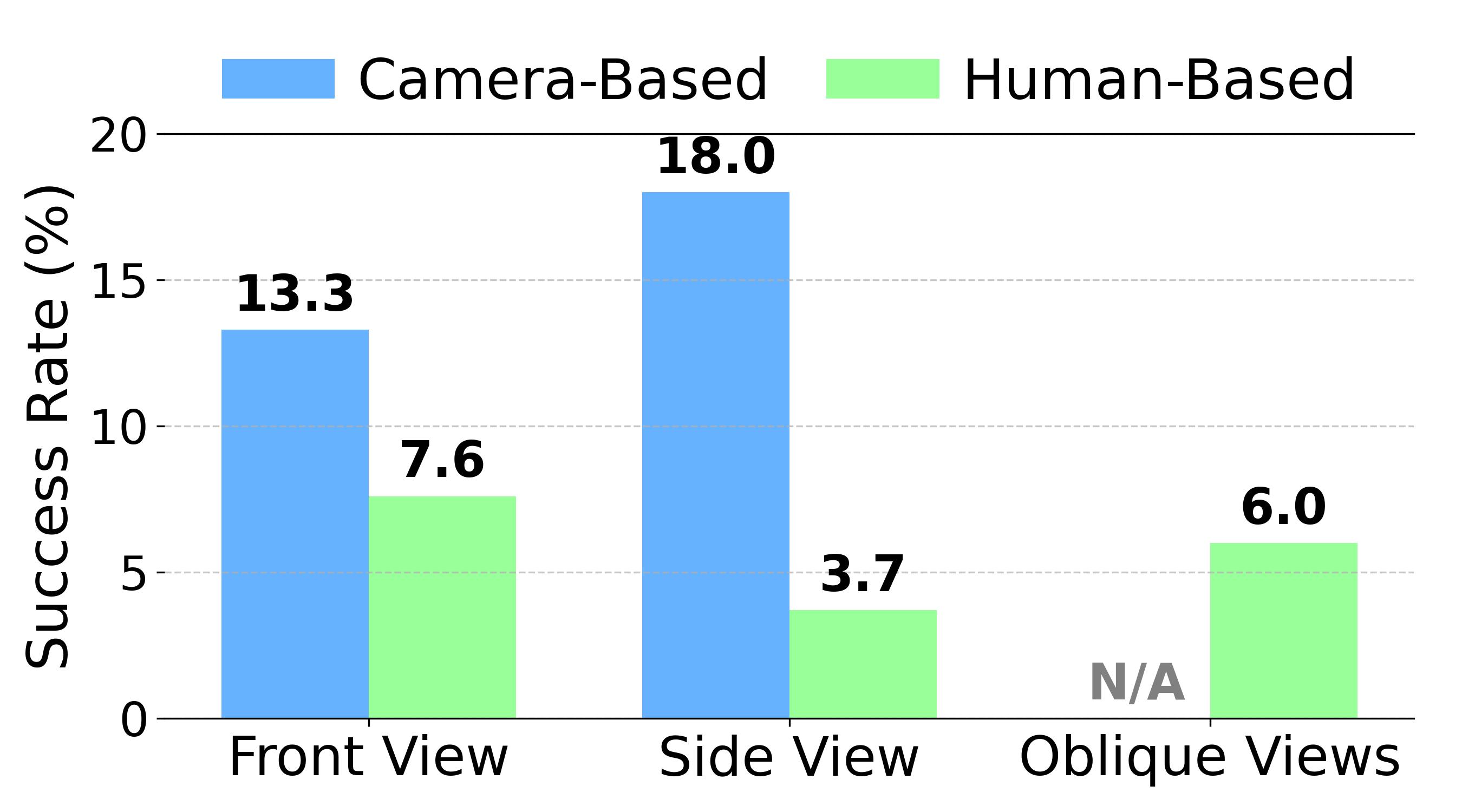}
        {\small (b) Qwen3-VL-Instruct}
    \end{minipage}
    \hfill
    \begin{minipage}[t]{0.32\textwidth}
        \centering
        \includegraphics[width=\linewidth]{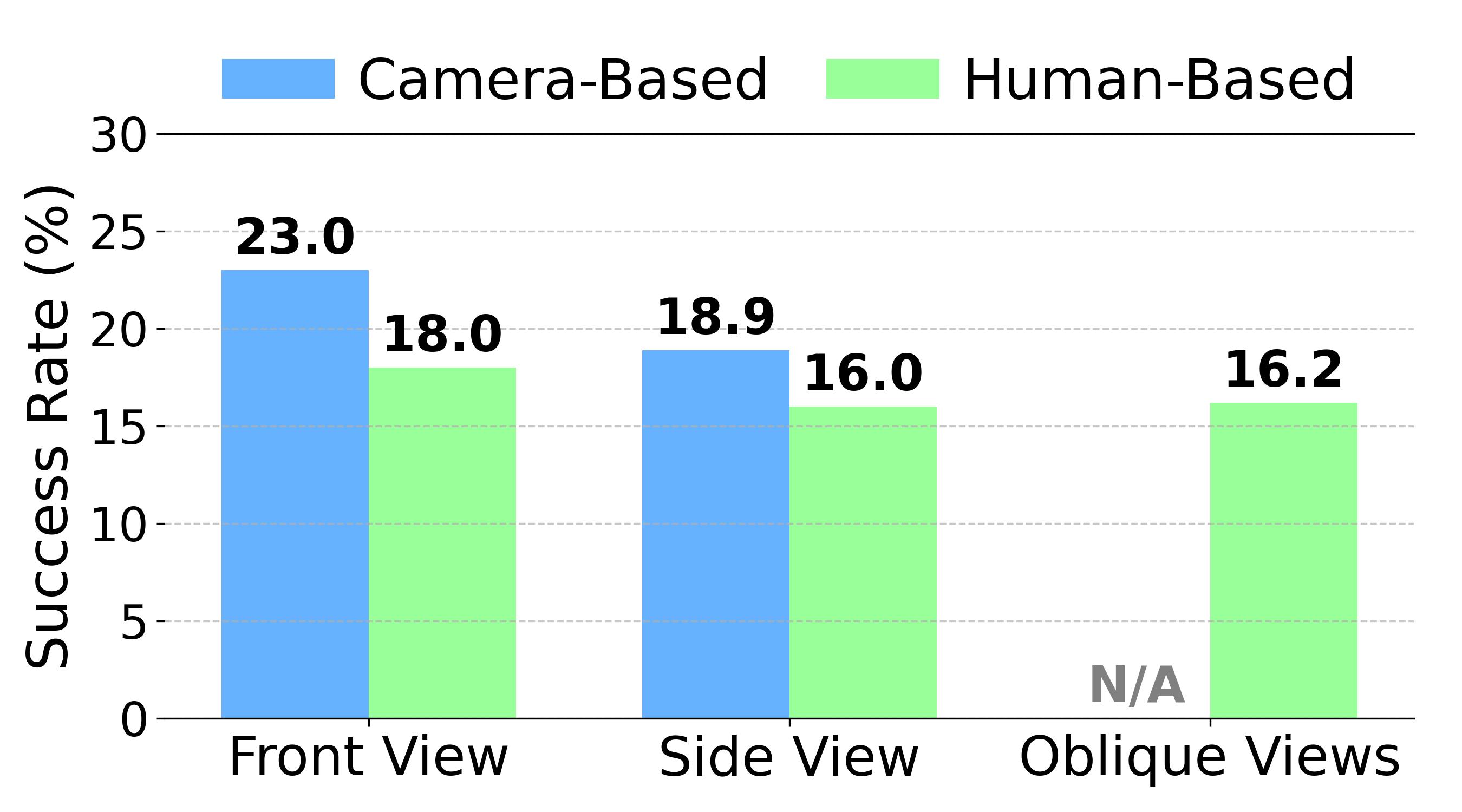}
        {\small (c) Qwen3-VL-Thinking}
    \end{minipage}
    \caption{\textbf{Spatial reasoning evaluation for VLMs.} VLM
performance across camera viewpoints and reference frames
(human-centered vs.\ camera-centered), defined in
Section~\ref{sec:spatial_design}. Camera-based references do not apply
under Oblique View, so only human-based references are reported there.\vspace{-0.3cm}}
    \label{fig:spatial_reasoning}
\end{figure*}

\noindent\textbf{Spatial Reasoning.}
\name varies two spatial conditions: the \emph{camera viewpoint} the
video was filmed from, and the \emph{reference frame} the instruction
uses, either human- or camera-centered. The reference frame has the
larger effect (Figure~\ref{fig:spatial_reasoning}). At fixed viewpoints, switching from a camera- to a human-centered frame lowers
GPT-5.4's Plan SR by 19.5 and 11.0 points, while camera viewpoint
matters little once the frame is human-centered (2.6 points between
side and front views).

\noindent\textbf{Generalization Analysis.}
Table~\ref{tab:indomain_vs_outdomain} evaluates LIBERO-finetuned policies on two splits of \name. The \emph{in-domain} split mirrors LIBERO's objects, layouts, and instructions, while the \emph{out-of-domain} split includes new instructions and newly 3D-scanned objects. Current policies generalize poorly. The best policy, $\pi_{0.5}$, reaches 47.1\% Task SR and 63.2\% PR in-domain but drops to 10.6\% Task SR and 20.0\% PR out-of-domain. LingBot-VA falls near zero out-of-domain. These results indicate that current policies struggle to handle unseen objects and instructions.

\vspace{-0.2cm}
\subsection{Real-World Experiments}
\label{sec:real_robot}
\vspace{-0.2cm}

\noindent\textbf{Setup.}
To test whether \name's simulation trends transfer to the real world,
we run experiments on a Franka Research 3 with a Robotiq-2F85 gripper,
following the DROID setup~\cite{khazatsky2024droid} with a ZED 2i
external and ZED Mini wrist camera. We fine-tune $\pi_{0.5}$ on 340
collected demonstrations of primitive actions such as \textit{pick}
and \textit{place}, and evaluate on 10 tasks (5 in-domain, 5
out-of-domain with unseen instructions or objects) at 5 trials each,
under both oracle-plan and VLM-plan execution.

\begin{table}[t]
\centering
\small
\renewcommand{\arraystretch}{1.1}
\setlength{\tabcolsep}{4pt}
\begin{minipage}[t]{0.48\textwidth}
    \centering
    \caption{\textbf{In-Domain vs.\ Out-of-Domain Execution.}
    Oracle-plan Task SR and PR for LIBERO-finetuned policies. Both
    degrade out-of-domain.}
    \label{tab:indomain_vs_outdomain}
    \vspace{0.3em}
    \begin{tabular}{l|cc|cc}
    \toprule
        Method
        & \multicolumn{2}{c|}{\cellcolor{yellow!10}In-Domain}
        & \multicolumn{2}{c}{\cellcolor{orange!10}Out-of-Domain} \\
        & Task SR & PR
        & Task SR & PR \\
        \midrule
        $\pi_{0.5}$   & \textbf{47.1} & \textbf{63.2} & \textbf{10.6} & \textbf{20.0} \\
        Lingbot-VA    & 1.4 & 2.8 & 0.7 & 2.6 \\
        \bottomrule
    \end{tabular}
\end{minipage}
\hfill
\begin{minipage}[t]{0.48\textwidth}
    \centering
    \caption{\textbf{Real-Robot Experiment.} $\pi_{0.5}$ on a Franka
    Research 3 under oracle and VLM-generated plans, in-domain
    vs.\ out-of-domain.}
    \label{tab:robotic_policy_categories}
    \vspace{0.3em}
    \begin{tabular}{l|cc|cc}
    \toprule
        Plan
        & \multicolumn{2}{c|}{\cellcolor{yellow!10}In-Domain}
        & \multicolumn{2}{c}{\cellcolor{orange!10}Out-of-Domain} \\
        & Task SR & PR
        & Task SR & PR \\
        \midrule
        Oracle Plan
        & \textbf{44.0} & \textbf{57.0}
        & \textbf{8.0} & \textbf{24.0} \\
        GPT-5.4
        & 24.0 & 47.0
        & 4.0 & 28.0 \\
        \bottomrule
    \end{tabular}
\end{minipage}
\end{table}

\noindent\textbf{Results.}
Table~\ref{tab:robotic_policy_categories} reports real-robot performance. $\pi_{0.5}$'s Task SR and PR drop sharply from in-domain to out-of-domain under both oracle and VLM-generated plans, mirroring the generalization gap observed in simulation. Because the same failure pattern appears on real hardware, the limitations \name surfaces reflect the capabilities of current systems rather than properties of the simulator.

\vspace{-0.2cm}
\section{Conclusion}
\label{sec:conclusion}
\vspace{-0.2cm}

We introduced \name, a benchmark for robot manipulation grounded in
observed human behavior, where a robot must reason about what a
person did in a scene to act on a language instruction. \name pairs
3,000 real-world human-action videos with aligned, executable LIBERO
tasks across 14 scenarios in four capability domains, and a
disentangled protocol separately measures video-to-plan reasoning and
low-level execution. Across nine VLMs and four policies, in
simulation and on a real robot, current systems remain far from
solving the benchmark. We hope \name serves as a
testbed for the reasoning and control capabilities that robots
working alongside people will require.

\textbf{Limitations.}
\name has several limitations that point to future work. First,
execution is evaluated in simulation, which abstracts away the contact
dynamics and perceptual noise of the physical world. Our real-robot
experiments reproduce the same failure patterns, suggesting the trends
transfer, but a full real-world benchmark at this scale remains future
work. Second, the human videos follow a filming guide, capturing
purposeful, well-segmented activity rather than fully unstructured
behavior. This control is what makes reliable video-simulator
alignment possible at scale, and relaxing it is a natural next step.
Third, the four capability domains are not exhaustive, and the
high-level action space is restricted to four primitives, bounding the
manipulation complexity each task can express.

\clearpage
\acknowledgments{This work was supported by Laboratory for Analytic Sciences via NC State University, ONR Award N00014-23-1-2356, Sony Focused Research award, and NSF CAREER Award 2541848.}

\bibliography{references/references}  %

\clearpage
\section*{Appendix}
\label{app:details}
\setcounter{subsection}{0}
\renewcommand{\thesubsection}{\arabic{subsection}}
\subsection{Benchmark Details}
\noindent{\textbf{Benchmark Statistics.}} \name comprises 3{,}000 instances across 14 tasks organized into four capability domains: 890 instances in Event Grounding, 725 in Procedural Reasoning, 455 in Implicit Intent Inference, and 930 in Episodic Reasoning. Each instance contains a language instruction, a real-world demonstration video, an aligned simulator task, and an oracle high-level action plan. In total, \name includes 3{,}000 videos spanning 20.1 hours, with an average length of 24.1 seconds each. Across all oracle plans, \name contains 12{,}690 steps in total, averaging 4.23 steps per task. For the simulator, we use objects from LIBERO and additionally 3D-scan 10 new real-world objects to build their corresponding digital twins.

\noindent{\textbf{Data Generation Prompts.}} We use a frontier LLM (e.g., GPT-5.1) to generate task specifications, prompting it with the shared object inventory, named regions, and primitive action space registered between the simulator and the real-world scene. The prompt is organized into three parts: it establishes the model's role and the required outputs (Part 1), specifies the environment and action space (Part 2), and defines the target task together with its constraints (Part 3). For each task category, we adapt Part 3 accordingly and pair the prompt with a few in-context examples that fix the output format. We show the prompt for the Imitation task below as a representative example.

\begin{tcolorbox}[
enhanced,
breakable,
title=\textbf{Part 1: Role, Context, \& Output Requirements},
label={box:part1},
colback=gray!2,
colframe=blue!35!black,
colbacktitle=blue!8,
coltitle=black,
boxrule=0.5pt,
arc=2pt,
left=8pt,
right=8pt,
top=10pt,
bottom=8pt,
fonttitle=\bfseries,
fontupper=\small,
attach boxed title to top left={xshift=3mm,yshift=-2mm},
boxed title style={
colback=blue!8,
colframe=blue!35!black,
boxrule=0.5pt,
arc=2pt,
left=5pt,
right=5pt,
top=2pt,
bottom=2pt
},
before skip=0.6em,
after skip=0.6em
]
\noindent{\color{blue!45!black}\textbf{Role.}}
You are an expert in robot learning and benchmark design for video-grounded manipulation tasks.
\vspace{0.5em}

\noindent{\color{blue!45!black}\textbf{Context.}}
We are building WatchAct, a benchmark where each instance pairs a real-world video of human tabletop activity with a language instruction and an aligned executable LIBERO simulation task. The goal is to evaluate whether a robot can reason over observed human behavior and act accordingly in simulation.
\vspace{0.5em}

\noindent{\color{blue!45!black}\textbf{Output.}}
For each instance, produce two paired outputs:
\begin{itemize}[leftmargin=1.3em, itemsep=2pt, topsep=3pt, parsep=0pt]
\item \textbf{Human Video Filming Instructions} --- a step-by-step recording guide, including the object list, the initial object layout, and the sequence of human actions.
\item \textbf{Robot Simulation Task Definition} --- the corresponding executable LIBERO task configuration, including a language instruction, a BDDL task definition, and an oracle high-level action plan.
\end{itemize}
\end{tcolorbox}

\begin{tcolorbox}[
enhanced,
breakable,
title=\textbf{Part 2: Environment \& Action Space},
label={box:part2},
colback=gray!2,
colframe=blue!35!black,
colbacktitle=blue!8,
coltitle=black,
boxrule=0.5pt,
arc=2pt,
left=8pt,
right=8pt,
top=10pt,
bottom=8pt,
fonttitle=\bfseries,
fontupper=\small,
attach boxed title to top left={xshift=3mm,yshift=-2mm},
boxed title style={
colback=blue!8,
colframe=blue!35!black,
boxrule=0.5pt,
arc=2pt,
left=5pt,
right=5pt,
top=2pt,
bottom=2pt
},
before skip=0.6em,
after skip=0.6em
]
\noindent{\color{blue!45!black}\textbf{Scene.}}
A tabletop environment shared between the real-world video and the simulation.
\vspace{0.5em}

\noindent{\color{blue!45!black}\textbf{Available elements.}}
\begin{itemize}[leftmargin=1.3em, itemsep=2pt, topsep=3pt, parsep=0pt]
\item \textbf{Objects:} around 30 tabletop objects, including food items, tableware, containers, and other household objects.
\item \textbf{Regions:} 11 tabletop regions and 6 container regions used to align real-world and simulated layouts.
\item \textbf{Predicates:} symbolic predicates used to define initial and goal states, such as \texttt{in}, \texttt{on}, \texttt{open}, and \texttt{closed}.
\end{itemize}
\vspace{0.5em}

\noindent{\color{blue!45!black}\textbf{Primitive Actions.}}
\texttt{PICK(object, region)}, \texttt{PLACE(object, region)}, \texttt{OPEN(object)}, \texttt{CLOSE(object)}.
\vspace{0.5em}

\noindent{\color{blue!45!black}\textbf{Task definition format.}}
Each simulation task is specified by a language instruction, the objects involved, an initial state, a goal state, and an oracle action plan. The real-world and simulated layouts should be aligned as closely as possible.
\end{tcolorbox}

\begin{tcolorbox}[
enhanced,
breakable,
title=\textbf{Part 3: Task Specification \& Constraints},
label={box:part3},
colback=gray!2,
colframe=blue!35!black,
colbacktitle=blue!8,
coltitle=black,
boxrule=0.5pt,
arc=2pt,
left=8pt,
right=8pt,
top=10pt,
bottom=8pt,
fonttitle=\bfseries,
fontupper=\small,
attach boxed title to top left={xshift=3mm,yshift=-2mm},
boxed title style={
colback=blue!8,
colframe=blue!35!black,
boxrule=0.5pt,
arc=2pt,
left=5pt,
right=5pt,
top=2pt,
bottom=2pt
},
before skip=0.6em,
after skip=0.6em
]
\noindent{\color{blue!45!black}\textbf{Task Category.}}
Design a task instance for the specified WatchAct category. The human video should provide the necessary context, and the robot task should require using both the video and the language instruction to infer the correct behavior.
\vspace{0.5em}

\noindent{\color{blue!45!black}\textbf{Note.}}
Imitation is used here as an example category. In Imitation, the robot reproduces the human action sequence. Other WatchAct categories may require different forms of reasoning, such as grounding an event, resolving an ambiguous reference, restoring a previous state, continuing an unfinished task, or correcting an error.
\vspace{0.5em}

\noindent{\color{blue!45!black}\textbf{Constraints.}}
\begin{itemize}[leftmargin=1.3em, itemsep=2pt, topsep=3pt, parsep=0pt]
\item Use only actions that map directly to the primitive action space; exclude non-discrete actions such as pouring, wiping, cutting, or stirring.
\item Select multiple objects to compose the real-world and simulated scenes.
\item Produce 5 candidate paired `Filming Instructions'' and `Task Definition'' entries for this generation batch, output in JSON format.
\item \textbf{Special requirement:} for each instance, include the intermediate subgoal state after each primitive action, so the temporal sequence can be evaluated step by step.
\end{itemize}
\vspace{0.5em}

\noindent{\color{blue!45!black}\textbf{In-context examples.}}
Complete worked examples are provided as output-format references.
\end{tcolorbox}

\subsection{Experimental Details}

\noindent{\textbf{Baselines.}} In this section, we introduce the four state-of-the-art robotic policies used in our experiments, along with their corresponding settings: three VLA models---\(\pi_{0.5}\)~\cite{intelligence2025pi_}, OpenVLA-OFT~\cite{kim2025fine}, and UniVLA~\cite{bu2025univla}---and one world-model-based policy, LingBot-VA~\cite{li2026causal}. We use publicly released LIBERO-finetuned checkpoints for all policies, applying them to \name with their default inference-time settings.

\begin{itemize}
    \item \(\pi_{0.5}\)~\cite{intelligence2025pi_}.
    \(\pi_{0.5}\) is a vision-language-action model designed for open-world generalization. It builds on a pretrained vision-language backbone and generates continuous low-level actions through a flow-matching action expert. We use the official openpi \texttt{pi05\_libero} checkpoint, fine-tuned on the combined LIBERO suites (LIBERO-Spatial, LIBERO-Object, LIBERO-Goal, and LIBERO-10).

    \item OpenVLA-OFT~\cite{kim2025fine}.
    OpenVLA-OFT is an optimized fine-tuning recipe for OpenVLA, using continuous action prediction, parallel decoding, action chunking, and an L1 regression objective to improve inference efficiency and execution performance. We use the authors' official LIBERO-finetuned checkpoint, specifically the LIBERO-Long version.

    \item UniVLA~\cite{bu2025univla}.
    UniVLA learns task-centric latent actions from large-scale videos, decoupling task intent from embodiment-specific control. A policy predicts latent actions from observations and instructions, and a lightweight decoder maps them to low-level robot commands. The latent action model is pretrained on manipulation, navigation, and human-video data. We use the LIBERO-Long checkpoint for long-horizon tasks with multiple sub-goals.

    \item LingBot-VA~\cite{li2026causal}.
    LingBot-VA is a world-model-based policy that combines video world modeling with robot action prediction. Rather than only mapping the current observation to an action, it jointly models future visual dynamics and action generation, providing temporal structure for long-horizon manipulation. We use the released LIBERO-Long checkpoint.
\end{itemize}

\noindent\textbf{Evaluation Protocol.} \name evaluates video-grounded embodied agents along three complementary axes: (i) the ability of VLMs to infer task plans from video and language; (ii) the ability of policies to execute oracle plans; and (iii) the ability of integrated pipelines to map visual perception and task reasoning into executable actions.

\begin{itemize}
    \item \textbf{Video-to-plan Reasoning.}
    In this setting, the model receives the video-language context \(c_i=(V_i,\ell_i)\) and predicts a high-level task plan:
    \[
        \hat{P}_i = f_{\theta}(c_i).
    \]
    We evaluate \(\hat{P}_i\) by checking it against the success predicate \(\phi_i\), and against the oracle plan \(P_i^\star\) when applicable. Specifically, the evaluation takes one of two forms, depending on the task structure. For order-sensitive tasks, where the oracle plan \(P_i^\star\) is unique, we compare \(\hat{P}_i\) directly against \(P_i^\star\) at the symbolic level. For tasks that admit multiple valid plans, we instead simulate \(\hat{P}_i\) with a rule-based symbolic executor and check whether the resulting final state satisfies \(\phi_i\). In both cases, the evaluation measures whether the predicted plan uses the correct objects, locations, spatial references, and temporal order.

    \item \textbf{Policy Execution with Oracle Plans.}
    The policy receives the robot observation \(o_{t}\) and the current oracle command \(u_{i,k}^\star\) from \(P_i^\star\), and outputs a low-level robot action:
    \[
        a_t \sim \pi_{\mathrm{exec}}(\cdot \mid o_{t}, u_{i,k}^\star).
    \]
    Commands \(u_{i,k}^\star\) are issued sequentially over fixed execution intervals. This setting evaluates whether the policy can carry out a correct high-level plan in \(\mathcal{T}_i\), with success determined by \(\phi_i\).

    \item \textbf{Integrated Planner and Policy.}
    In this setting, the agent receives the video-language context \(c_i=(V_i,\ell_i)\) and the robot observation \(o_{t}\), and directly outputs low-level robot actions:
    \[
        a_t \sim \pi_{\mathrm{e2e}}(\cdot \mid c_i, o_{t}).
    \]
    It is evaluated by whether the robot satisfies \(\phi_i\) before the episode times out.
\end{itemize}

\subsection{Additional Results}

\noindent{\textbf{Main Results.}}
We report the main results for nine VLMs on video-to-plan reasoning in Table~\ref{tab:appendix_subtask_results_vlm}, including models not covered in the main paper's Table~\ref{tab:subtask_results_vlm} (Gemini-3.1-Flash-Lite, GPT-5.1, and Gemini-3-Flash), broken down by the fourteen task categories (Table~\ref{tab:taxonomy_brief}).

\begin{table}[hbt]
   \centering
   \caption{\textbf{Main Results on Video-to-plan Reasoning.}
   Each predicted VLM plan is run through a symbolic executor and scored using the Plan SR metric, which measures whether the resulting object states satisfy the task's success conditions. Even the strongest model, Gemini-3.1-Pro, reaches only 36.8\% average Plan SR, 60.3 points below human performance.
   Task abbreviations: \textbf{Act}: Fine-Grained Action; \textbf{Cnt}: Count;
   \textbf{Ord}: Ordinal; \textbf{SC}: State Change; \textbf{Mmt}: Moment;
   \textbf{Imit}: Imitation; \textbf{Rev}: Reversal; \textbf{TS}: Temporal
   Sort; \textbf{NC}: Nonverbal Cue; \textbf{RD}: Reference
   Disambiguation; \textbf{Rest}: Restore; \textbf{TC}: Task Continuation;
   \textbf{EC}: Error Correction; \textbf{Cond}: Conditional Execution.}
   \label{tab:appendix_subtask_results_vlm}
      \vspace{0.5em}
      \renewcommand{\arraystretch}{1.3}
      \setlength{\tabcolsep}{2pt}
      \resizebox{\textwidth}{!}{
          \begin{tabular}{l|cccccc|cccc|ccc|ccccc|c}
          \toprule
          \multirow{3}{*}{Methods} &
          \multicolumn{6}{c}{\cellcolor{yellow!10}Event} &
          \multicolumn{4}{c}{\cellcolor{orange!10}Procedural} &
          \multicolumn{3}{c}{\cellcolor{pink!20}Implicit Intent} &
          \multicolumn{5}{c}{\cellcolor{brown!10}Episodic} &
          \multirow{3}{*}{Overall} \\
           &
          \multicolumn{6}{c}{\cellcolor{yellow!10}Grounding} &
          \multicolumn{4}{c}{\cellcolor{orange!10}Reasoning} &
          \multicolumn{3}{c}{\cellcolor{pink!20}Inference} &
          \multicolumn{5}{c}{\cellcolor{brown!10}Reasoning} & \\
          \cmidrule(lr){2-7} \cmidrule(lr){8-11} \cmidrule(lr){12-14} \cmidrule(lr){15-19}
           &
          Act & Cnt & Ord & SC & Mmt & \textbf{Avg.} &
          Imit & Rev & TS & \textbf{Avg.} &
          NC & RD & \textbf{Avg.} &
          Rest & TC & EC & Cond & \textbf{Avg.} & \\
          \midrule
           Human Performance
          & 100.0 & 93.3 & 93.3 & 100.0 & 100.0 & 97.3
          & 96.7 & 96.7 & 96.7 & 96.7
          & 93.3 & 96.7 & 95.0
          & 100.0 & 96.7 & 93.3 & 100.0 & 98.0
          & 97.1 \\
          \midrule
          \rowcolor{navyblue!5} \multicolumn{1}{l|}{\textcolor{black}{\textbf{\textit{Open-source Models}}}} & & & & & & & & & & & & & & & & & & & \\
          Qwen3-VL-235B-Instruct
          & 4.9 & 10.6 & 3.7 & 14.8 & 5.9 & 7.5
          & 12.8 & 8.4 & 8.6 & 10.2
          & 6.7 & 4.2 & 5.3
          & 5.0 & 27.6 & 22.4 & 18.3 & 17.0
          & 10.8 \\
          InternVL3.5-241B
          & 16.9 & 15.0 & 12.1 & 21.3 & 0.8 & 13.8
          & 7.3 & 4.0 & 4.7 & 5.5
          & 2.6 & 10.4 & 7.0
          & 4.2 & 33.5 & 30.4 & 22.7 & 21.0
          & 13.0 \\
          Qwen3-VL-235B-Thinking
          & 18.2 & 21.9 & 26.0 & 31.0 & 12.1 & 22.1
          & 9.7 & 6.7 & 7.2 & 8.0
          & 6.2 & 17.3 & 12.6
          & 10.8 & 44.1 & 36.0 & 30.3 & 28.3
          & 18.4 \\
          \midrule
          \rowcolor{navyblue!5} \multicolumn{1}{l|}{\textcolor{black}{\textbf{\textit{Proprietary Models}}}} & & & & & & & & & & & & & & & & & & & \\
   
   Gemini-3.1-Flash-Lite
   & 8.0 & 11.9 & 7.4 & 14.2 & 4.6 & 9.1
   & 9.0 & 4.9 & 6.1 & 6.9
   & 5.1 & 7.3 & 6.4
   & 3.1 & 42.9 & 34.4 & 21.1 & 23.0
   & 12.5 \\
   
   GPT-5.1
   & 30.2 & 26.2 & 27.4 & 54.6 & 7.4 & 29.6
   & 33.4 & 22.2 & 26.5 & 27.9
   & 4.6 & 18.1 & 12.4
   & 34.2 & 37.1 & 29.6 & 30.2 & 32.7
   & 27.4 \\
   
   Gemini-3-Flash
   & 29.3 & 40.6 & 40.9 & 53.5 & 23.1 & 37.5
   & 24.5 & 21.3 & 22.2 & 22.8
   & 23.1 & 31.2 & 27.8
   & 29.2 & 42.9 & 41.6 & 33.1 & 35.9
   & 31.9 \\
          
          GPT-5.4
          & \textbf{36.4} & 24.4 & 33.0 & \textbf{60.6} & 4.5 & 32.9
          & \textbf{32.8} & 27.1 & 27.6 & \textbf{29.6}
          & \textbf{21.0} & 20.4 & 20.7
          & \textbf{36.9} & 43.5 & 37.6 & 39.3 & 39.1
          & 31.5 \\
          Gemini-Robotics-ER-1.6
          & 33.3 & \textbf{40.6} & 44.7 & 45.2 & 19.2 & 37.4
          & 23.1 & 25.3 & 22.2 & 23.5
          & 17.9 & 28.5 & 24.0
          & 33.5 & 48.2 & 37.6 & 39.8 & 38.9
          & 32.4 \\
          Gemini-3.1-Pro
          & 33.8 & 36.9 & \textbf{47.4} & 60.0 & \textbf{22.7} & \textbf{40.7}
          & 30.0 & \textbf{30.2} & \textbf{28.1} & 29.5
          & 18.0 & \textbf{43.1} & \textbf{32.4}
          & 31.2 & \textbf{52.4} & \textbf{43.2} & \textbf{42.3} & \textbf{41.0}
          & \textbf{36.8} \\
       \bottomrule
      \end{tabular}}
   \end{table}

\noindent{\textbf{Effect of input FPS.}} We explore the effect of FPS sampling in the video-to-plan reasoning experiments. In Table~\ref{tab:fps}, we sample the open-source model Qwen3-VL-235B-Instruct and GPT-5.4 at 1, 4, and 8 FPS. The results show that increasing the sampling rate yields only modest gains. Both Qwen3-VL-Instruct and GPT-5.4 achieve their best results at 4 FPS (12.3\% Plan SR and 34.5\% Plan SR, respectively), only a small improvement over the 1 FPS setting and still far behind human 97.1\% Plan SR. This suggests that state-of-the-art models struggle with video-grounded embodied planning, and that this cannot be resolved by simply increasing the frame sampling rate.

\begin{table}[htb!]
\renewcommand{\arraystretch}{1.2}
\caption{\textbf{Effect of Input FPS.} We report the Plan SR of Qwen3-VL-235B-Instruct and GPT-5.4 under different FPS sampling rates. The results show that increasing the sampling rate yields only modest gains. GPT-5.4 achieves its highest score at 4 FPS (34.5\%), but still falls far behind the human Plan SR of 97.1\%, indicating that the video-to-plan task in WatchAct cannot be solved by simply increasing the sampling rate.}
\centering
\small
\setlength{\tabcolsep}{8pt}
\begin{tabular}{lccc}
\toprule
Model & 1 FPS & 4 FPS & 8 FPS \\
\midrule
Human & 97.1 & 97.1 & 97.1 \\
\midrule
Qwen3-VL-235B-Instruct & 10.8 & 12.3 & 11.5 \\
GPT-5.4 & \textbf{31.5} & \textbf{34.5} & \textbf{34.2} \\
\bottomrule
\end{tabular}
\label{tab:fps}
\end{table}

\subsection{Qualitative Examples}
\begin{figure}[p]
    \centering
    \includegraphics[width=\linewidth]{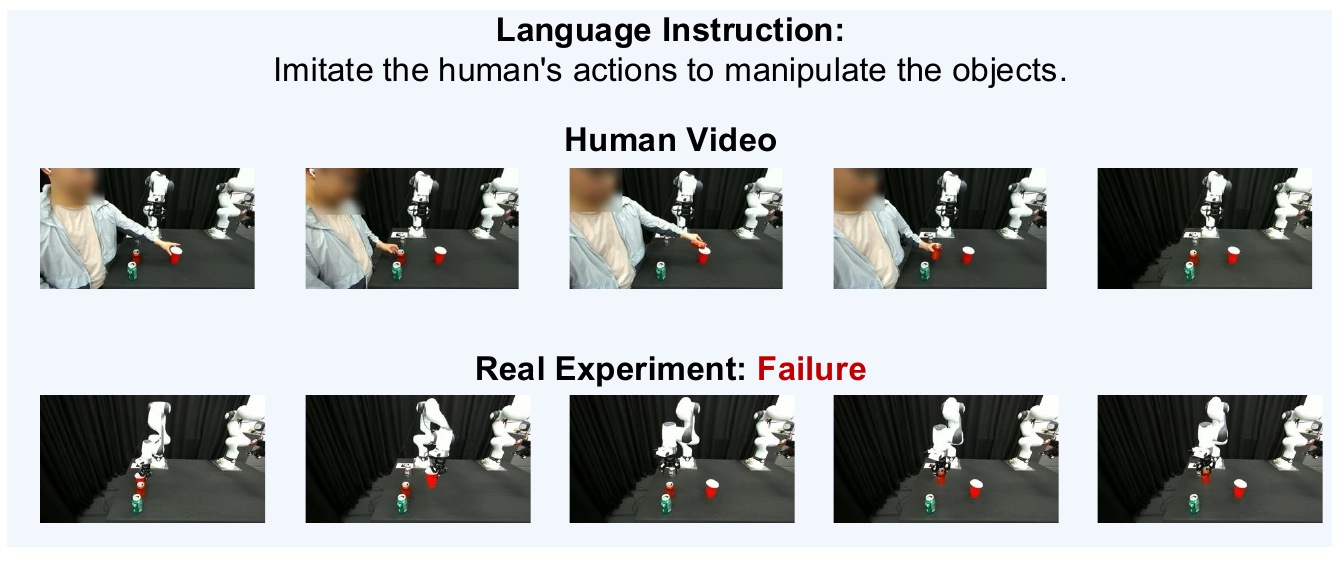}
    \caption{Real-world task example: Imitation. In the human video, the person moves a cup and pours in cola. The robot is asked to imitate this action. After picking up the cola, however, the robot fails at the pouring step.}
    \label{fig:appendix-real-world-4}
\end{figure}

\begin{figure}[p]
    \centering
    \includegraphics[width=\linewidth]{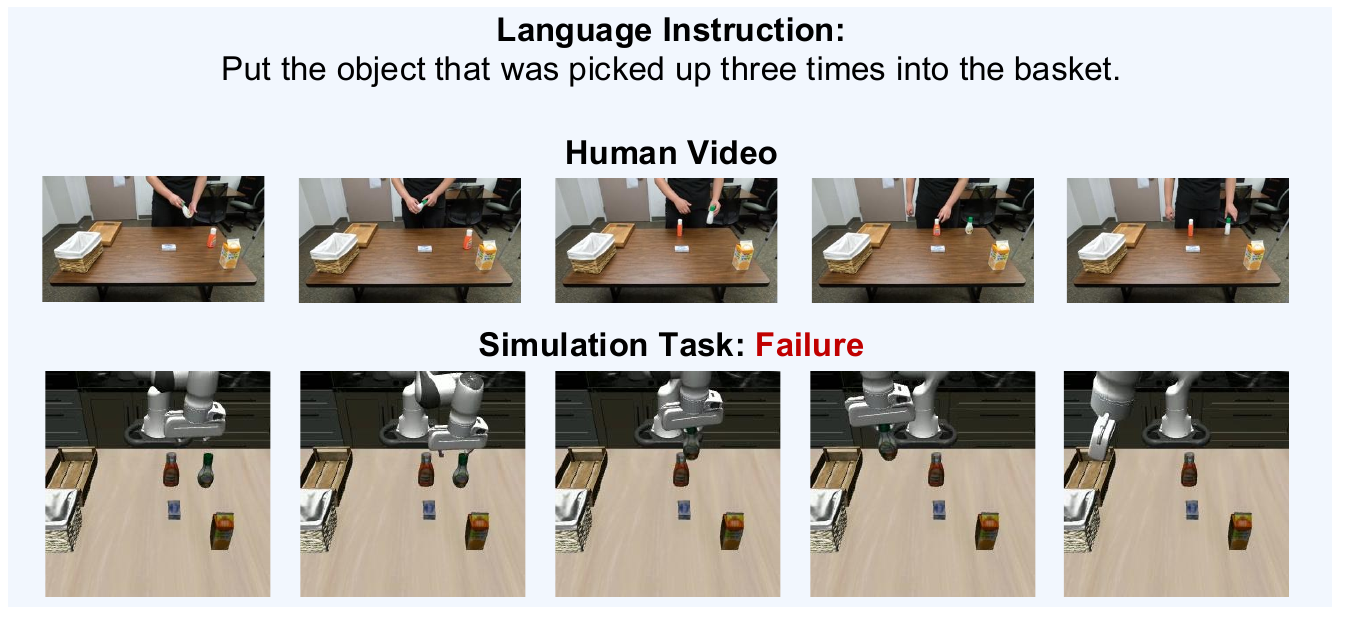}
    \caption{Simulation task: Count. In the human video, the person picks up the white bottle three times. Based on the number of actions, the robot must localize the white bottle and move it to the basket; the robot then moves it to the wooden tray.}
    \label{fig:appendix-count}
\end{figure}

\begin{figure}[p]
    \centering
    \includegraphics[width=\linewidth]{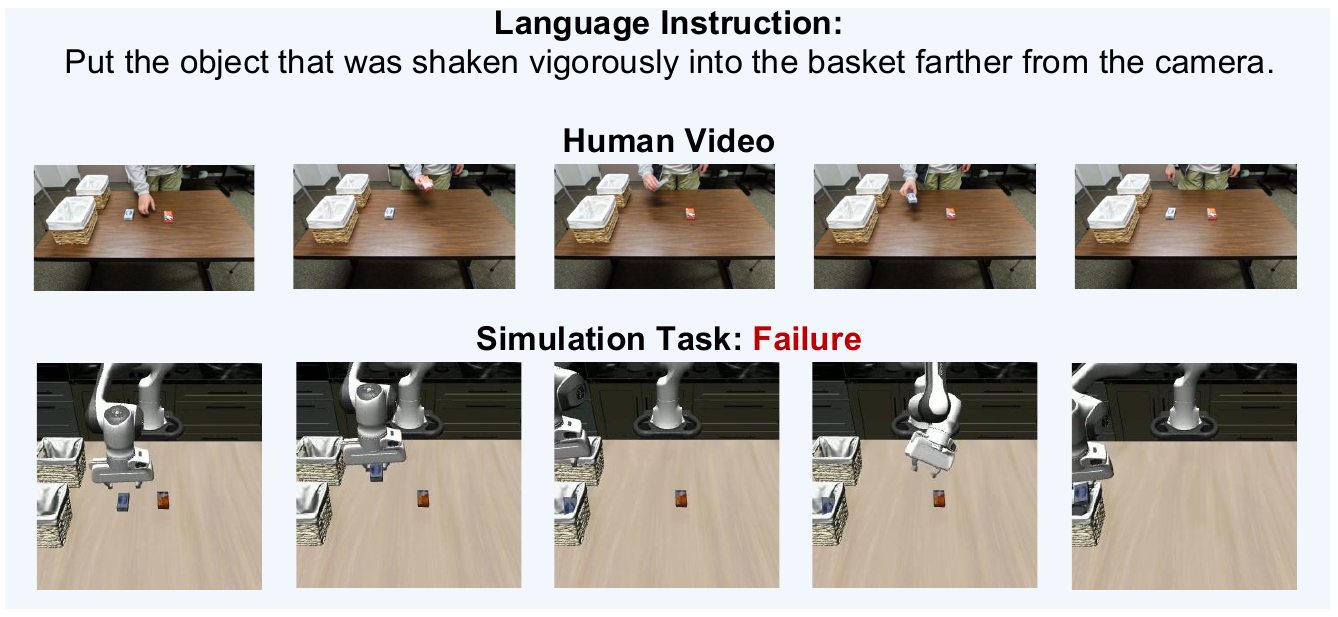}
    \caption{Simulation task: Fine-Grained Action. In the human video, the person vigorously shakes the blue box and gently picks up the red box. The robot only needs to move the blue box to the basket farther from the camera; however, after picking up the blue box, the robot mistakenly moves it to the other basket.}
    \label{fig:appendix-fine-grained-action}
\end{figure}

\begin{figure}[p]
    \centering
    \includegraphics[width=\linewidth]{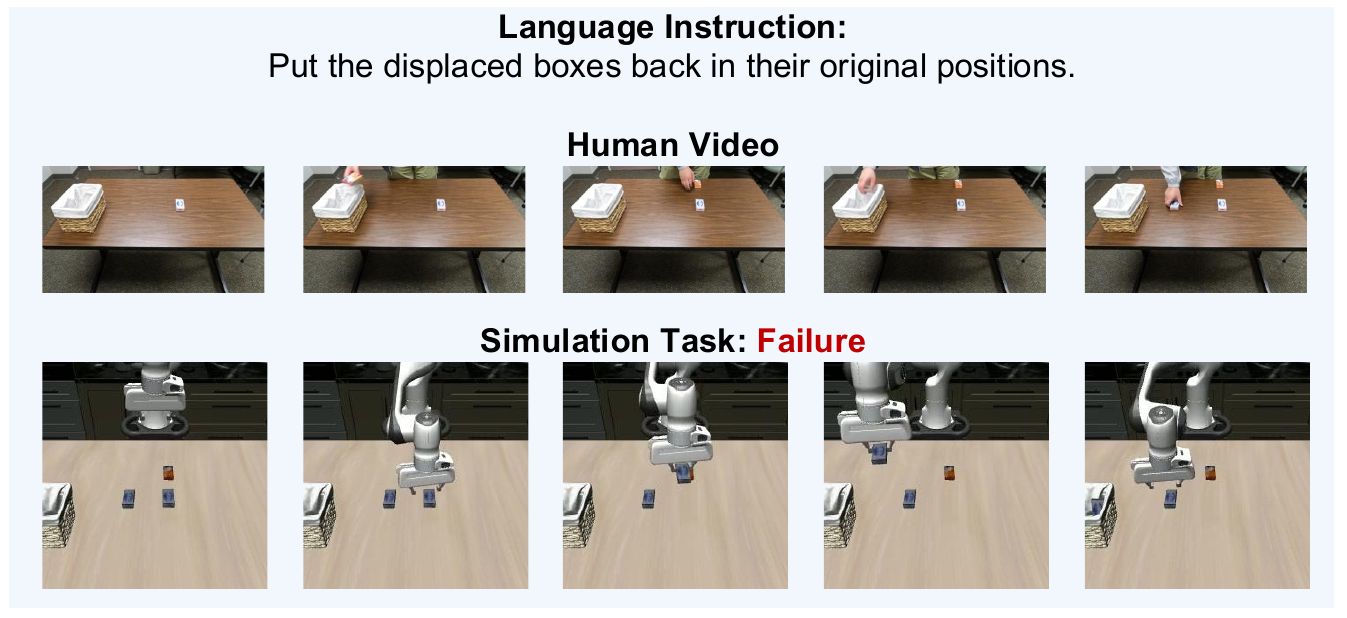}
    \caption{Simulation task: Restore previous state. In the human video, the person removes several objects from the basket and places them on the table. The robot is asked to return the displaced objects to their original positions; however, the robot mistakenly places an object that had not been moved into the basket.}
    \label{fig:appendix-previous}
\end{figure}

\begin{figure}[p]
    \centering
    \includegraphics[width=\linewidth]{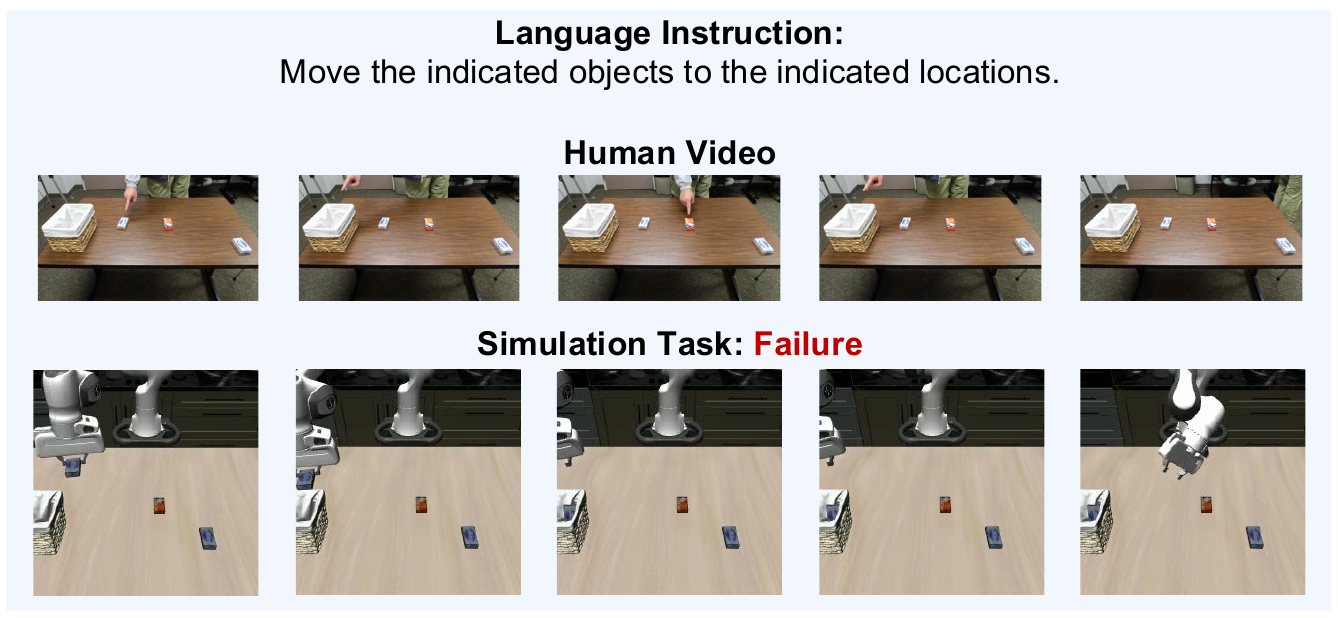}
    \caption{Simulation task: Nonverbal. In the human video, the person points to an object and then to the basket, signaling that the object should be placed in the basket. The robot is asked to move the object according to the person's intent; however, it fails to pick up the red box.}
    \label{fig:appendix-nonverbal-2}
\end{figure}

Qualitative Examples. In this section, we present and analyze concrete examples from both the real robot and the simulation platform. For each task, we feed the Oracle Plan to the policy and use \(\pi_{0.5}\) in both real-world and simulated experiments. We find that the robot exhibits two common failure modes under the Oracle Plan: (1) Long-horizon tasks are difficult. As shown in the examples in Figures~\ref{fig:appendix-real-world-4} and~\ref{fig:appendix-nonverbal-2}, long-horizon tasks require the model to succeed at every step, which poses a significant challenge. Moreover, we observe that on some tasks, when the input switches to the next step, the arm often fails to adjust to a suitable state, causing the next step to fail. (2) The policy follows instructions poorly. During execution, the policy may pick the wrong object (Figures~\ref{fig:appendix-previous},) or place an object in the wrong location (Figures~\ref{fig:appendix-fine-grained-action} and~\ref{fig:appendix-count}). This observation is consistent with prior work~\cite{fei2025libero,zhou2025libero} and is often caused by strong biases in the training data, which lead the model to rely on visual bias rather than truly understanding the instruction and reasoning correctly.
\begin{figure}[p]
    \centering
    \includegraphics[width=\linewidth]{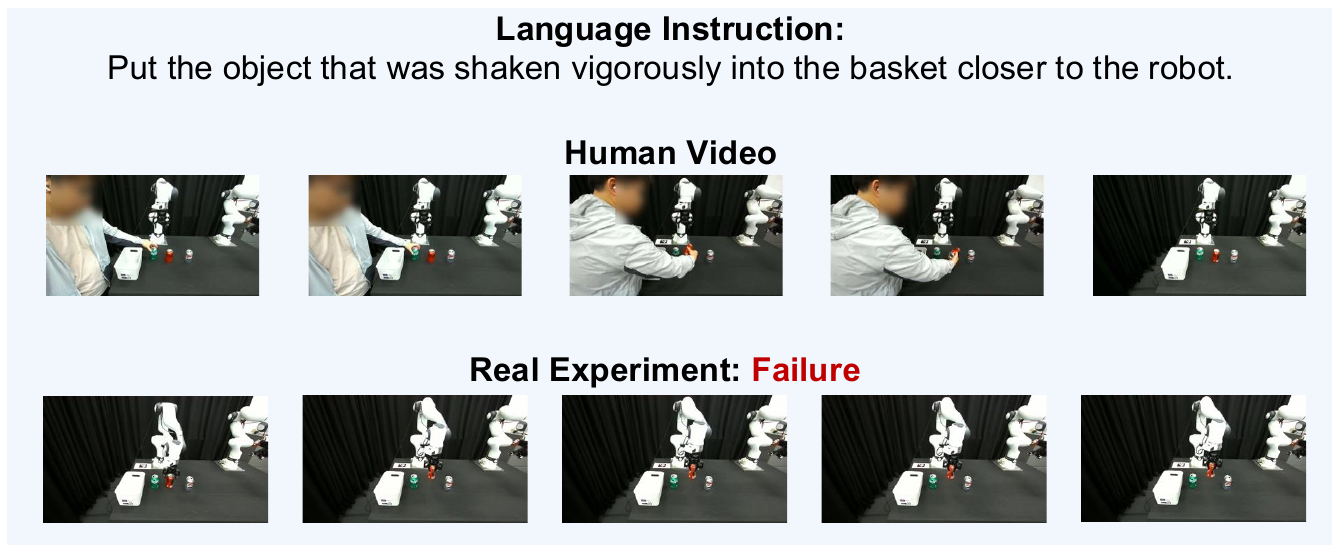}
    \caption{Real-world task example: Fine-Grained Action. In the human video, the person vigorously shakes the cola can. The robot must recognize the shaking action, localize the cola can, and follow the instruction to pick it up and place it into the basket. However, the robot fails during execution: after picking up the cola can, it fails to put it into the basket.}
    \label{fig:appendix-real-world-1}
\end{figure}

\begin{figure}[p]
    \centering
    \includegraphics[width=\linewidth]{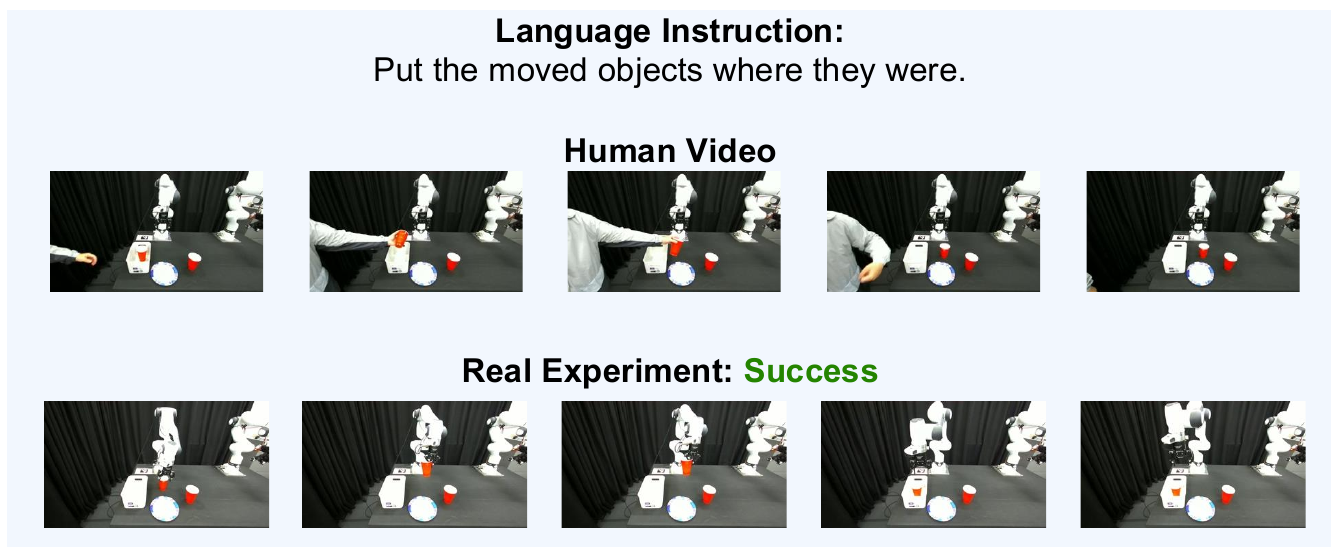}
    \caption{Real-world task example: Restore Previous State. In the human video, the person removes a cup from the basket. Following the instruction, the robot must successfully return the displaced cup to its original position.}
    \label{fig:appendix-real-world-2}
\end{figure}

\begin{figure}[p]
    \centering
    \includegraphics[width=\linewidth]{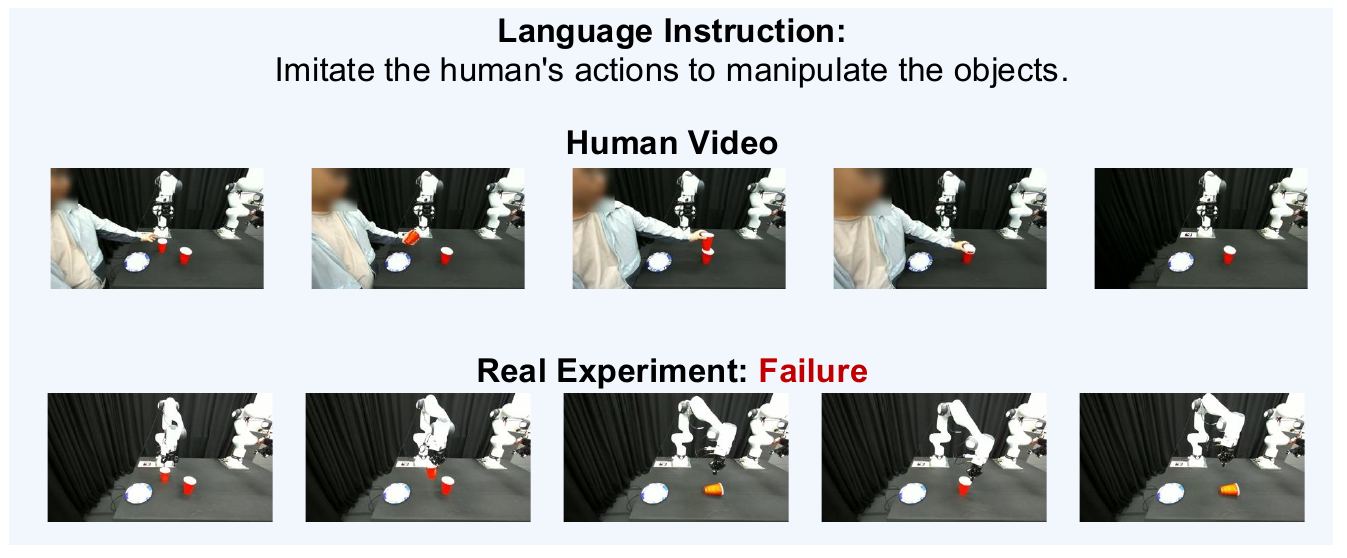}
    \caption{Real-world task example: Imitation. In the human video, the person picks up the cup on the left and places it into another cup. The robot is asked to imitate this action. Following the instruction, the robot places the cup into the other cup, but knocks it over, failing to keep its motion stable during execution.}
    \label{fig:appendix-real-world-3}
\end{figure}

\begin{figure}[p]
    \centering
    \includegraphics[width=\linewidth]{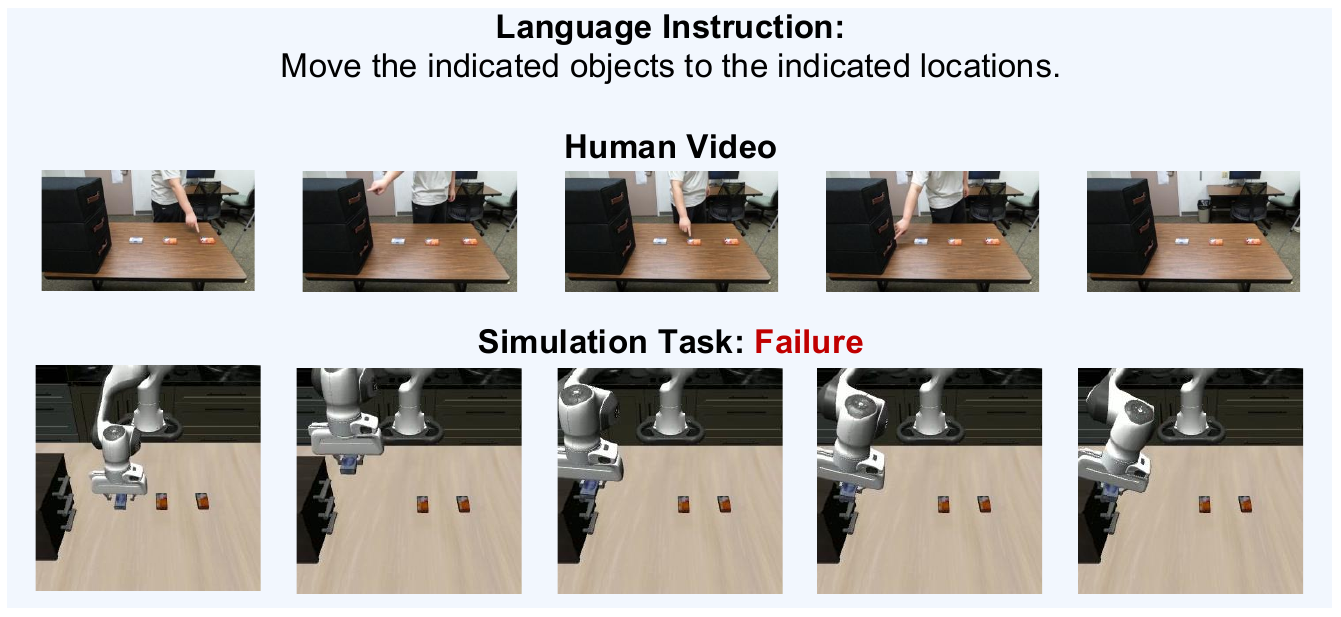}
    \caption{Simulation task: Nonverbal. In the human video, the person points to an object and then to a specific drawer, signaling that the object should be placed in the corresponding drawer. The robot is asked to move the object according to the person's gestures; however, the robot fails to open the drawer.}
    \label{fig:appendix-nonverbal-1}
\end{figure}

\begin{figure}[p]
    \centering
    \includegraphics[width=\linewidth]{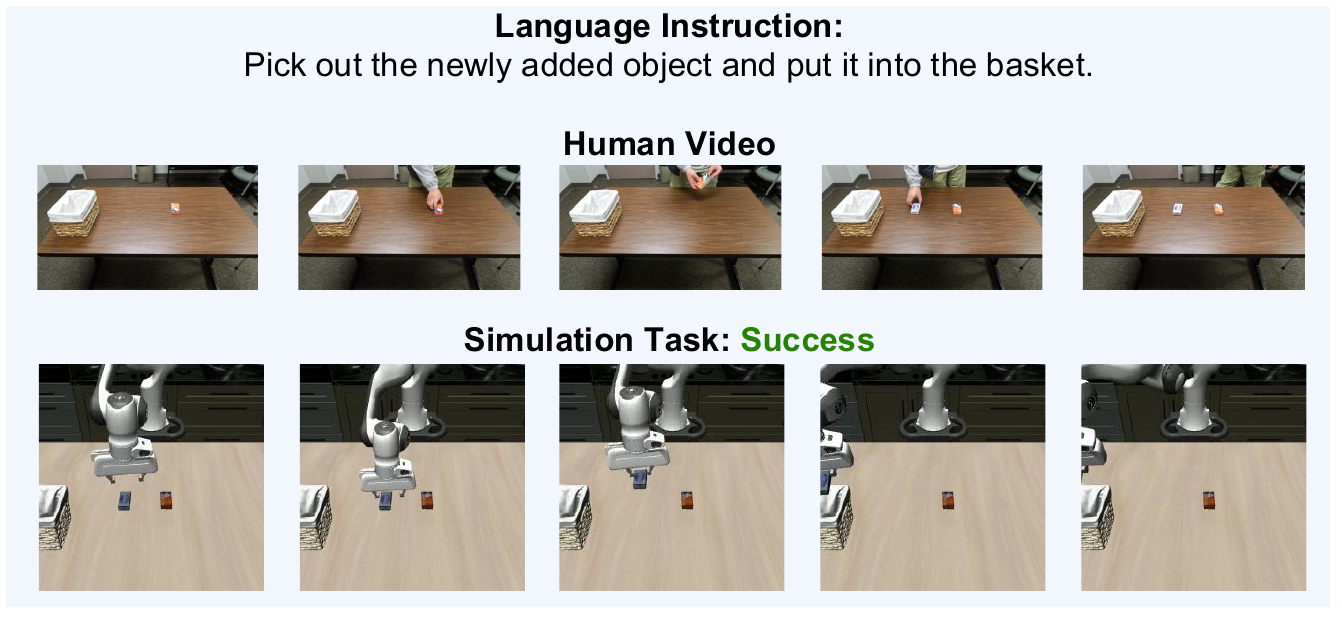}
    \caption{Simulation task: State change. In the human video, the person adds a new object to the table and asks the robot to place the newly added object into the basket. The robot identifies the new object as the blue box and successfully places it into the basket.}
    \label{fig:appendix-state-change}
\end{figure}

\end{document}